\documentclass{nle}
\usepackage{multirow}
\usepackage{color,soul}
\usepackage{algorithm}
\usepackage[noend]{algpseudocode}
\usepackage{comment}
\usepackage{adjustbox}
\usepackage{amsmath}
\title[Natural Language Engineering]
     {Cluster-based Mention Typing for Named Entity Disambiguation}
\author[Arda \c{C}elebi and Arzucan \"{O}zg\"{u}r]
       {Arda \c{C}elebi and Arzucan \"{O}zg\"{u}r\\
        Department of Computer Engineering \\
        Bo\u{g}azi\c{c}i University \\
        Bebek, 34342 \. {I}stanbul, Turkey \\
        {\tt \{ arda.celebi, arzucan.ozgur \} @boun.edu.tr} \\
       }

\DeclareMathOperator*{\argmin}{\textit{argmin}}
\pagerange{\pageref{firstpage}--\pageref{lastpage}}
\pubyear{2020}

\begin{document}

\label{firstpage}
\maketitle

\begin{abstract}

An entity mention in text such as ``Washington'' may correspond to many different named entities such as the city ``Washington D.C.'' or the newspaper ``Washington Post.'' The goal of named entity disambiguation is to identify the mentioned named entity correctly among all possible candidates. If the type (e.g. location or person) of a mentioned entity can be correctly predicted from the context, it may increase the chance of selecting the right candidate by assigning low probability to the unlikely ones. This paper proposes cluster-based mention typing for named entity disambiguation. The aim of mention typing is to predict the type of a given mention based on its context. Generally, manually curated type taxonomies such as Wikipedia categories are used. We introduce cluster-based mention typing, where named entities are clustered based on their contextual similarities and the cluster ids are assigned as types. The hyperlinked mentions and their context in Wikipedia are used in order to obtain these cluster-based types. Then,  mention typing models are trained on these mentions, which have been labeled with their cluster-based types through distant supervision. At the named entity disambiguation phase, first the cluster-based types of a given mention are predicted and then, these types are used as features in a ranking model to select the best entity among the candidates. We represent entities at multiple contextual levels and obtain different clusterings (and thus typing models) based on each level. As each clustering breaks the entity space differently, mention typing based on each clustering discriminates the mention differently. When predictions from all typing models are used together, our system achieves better or comparable results based on randomization tests with respect to the state-of-the-art levels on four defacto test sets.
\end{abstract}

\section{Introduction}

\begin{figure}[h]
\centering
\includegraphics[width=120mm]{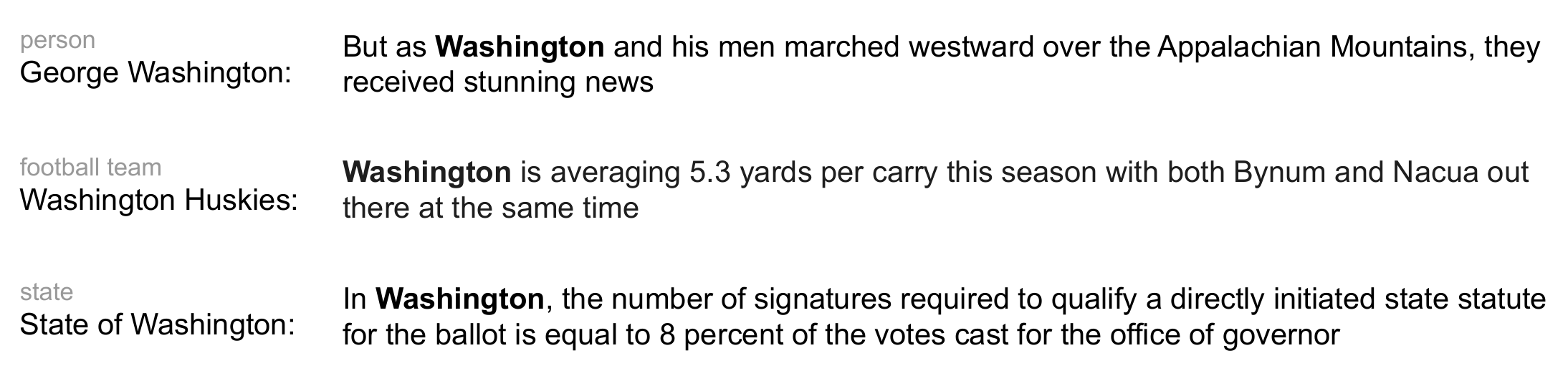}
\caption{Mentions of different named entities with the same surface form ``Washington.''}
\label{fig:WashingtonMentions}
\end{figure}

We humans name things around us in order to refer to them. Many times, different things can have the same name. For example, when you visit the disambiguation page\footnote{Visit https://simple.wikipedia.org/wiki/Washington\_(disambiguation)} for the word ``Washington'' in Wikipedia, you see tens of cities and counties in the United States that are named ``Washington.'' Moreover, the same word is used as a hyperlink title to refer to several different articles throughout Wikipedia. Things that can be denoted with a proper name are called named entities. The Knowledge Base (KB) of a machine keeps the list of all known named entities to that machine. When they are mentioned in a document, the task of identifying the correct named entity that a mention refers to among all the possible entities in the KB is called Named Entity Disambiguation (NED). Figure~\ref{fig:WashingtonMentions} gives three example sentences with the mention ``Washington'', each referring to a different named entity. For a human, it is not hard to figure out which entity is being referred to by considering the clues in the surrounding context of the mention. However, from a machine point of view, each mention may refer to any one of the hundreds of named entities in its KB.

NED, in general, is done in two steps. In the first step, the candidate named entities in the KB are identified  based on their lexical similarities to the given entity mention. In the second step, which is the actual disambiguation step, each candidate is scored based on some extracted features. The one with the highest score is returned as the predicted named entity corresponding to the input mention. In this field, the reference KB is most commonly based on Wikipedia, as is in our case.

In the literature, various approaches have been proposed to solve the NED task. Early studies used entity-specific local features like the similarity of the candidate to the document topic in order to score them individually \cite{Bunescu06,Mihalcea07}. Cucerzan \shortcite{Cucerzan07} proposed the idea that entities in a document should be correlated with each other and consistent with the topic of that document. How well a named entity is connected to the surrounding named entities is measured by coherence. They tried to optimize the coherence by including global context-based features that take into account the surrounding candidates into the disambiguation decision. Later studies looked at collective disambiguation, where all candidates are considered together in the decision process, rather than individually \cite{Kulkarni09}. Most of the collective models employed computationally complex graph-based approaches in order to find the sub-graph of candidates with the highest coherence. As the deep learning approaches advanced, Long Short-term Memory (LSTM) \cite{Phan17} models have been used to capture the long-range relations between words in the context, and attention-based neural networks have been used \cite{Ganea17} to pay more attention to the most relevant segments in the surrounding context of mentions. Entity embeddings, which are continuous vector-based representations of named entities, have been optimized to detect the relations between the candidate named entities and their context \cite{Yamada16}. A number of recent studies have investigated utilising the category hierarchy of Wikipedia for Named Entity Disambiguation. Raiman and Raiman \shortcite{Raiman18} proposed to integrate the symbolic structure of the category hierarchy of the named entities in Wikipedia in order to constrain the output of a neural network model. Murty \textit{et al.} \shortcite{Murty18} used the same principle of making use of type hierarchy, but proposed a hierarchically-aware training loss. Onoe and Durrett \shortcite{Onoe19} followed a similar approach, but rather than predicting the entities directly, they only modeled the fine-grained entity properties, which represent detailed entity type information. Then, they used that as a filtering mechanism at the disambiguation step. In addition to all these new techniques, more and more studies have been using  Wikipedia as a vast resource for representation learning from its text and the hyperlinked mentions in the text.

\begin{figure}[h]
\centering
\includegraphics[width=125mm]{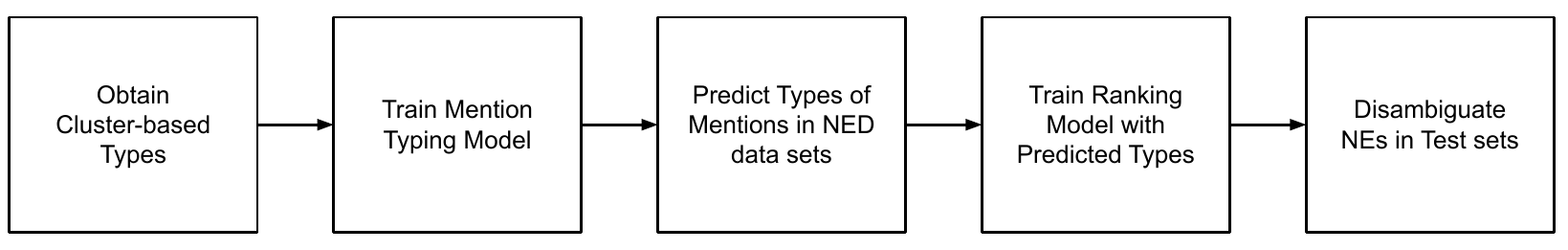}
\caption{The Workflow of the Proposed System (NE: Named Entity, NED: Named Entity Disambiguation).}
\label{fig:OverallFlow}
\end{figure}

This research focuses on identifying the type of a mentioned named entity first and then using this information to improve the prediction of the identity of that named entity at the disambiguation step. The first step is called mention-level entity typing, or mention typing in short. It is the task of predicting the type (e.g. politician, city etc.) of a single mention given its surface form and its surrounding context.
Prior studies on this task use manually curated type taxonomies, like Wikipedia categories, and assign each named entity mention to one or more such types. However, considering there are millions of named entities, such manual curation is inherently incomplete. Hence, we propose to obtain types automatically in an unsupervised fashion and use these types to improve NED. The workflow of the proposed approach is summarized in  Figure~\ref{fig:OverallFlow}. In the first step, we cluster named entities based on their contextual similarities in Wikipedia and use the cluster ids as types, hence cluster-based types. These types no longer correspond to conventional discrete types as exemplified in Figure~\ref{fig:WashingtonMentions}. Instead, each cluster-based type represents a cluster of named entities that are mentioned in similar contexts. Since the entities with the same conventional type tend to occur in similar context, it is also likely to obtain clusters that implicitly correspond to these types, like person or football team. In the second step, we train a mention typing model on the hyperlinked mentions in Wikipedia, which have been labeled with their cluster-based types through distant supervision. In the third step, mentions in the NED data sets are classified with this typing model. The fourth step involves using those type predictions as features for training an entity candidate ranking model. In the final step, for each entity mention in the NED test sets, the trained ranking model selects the most possible entity candidate in the KB. We use a simple feed-forward neural network model to score the candidates based on local context-independent and global context-based features, which include the aforementioned predictions. Moreover, in order to maximize the contribution from the context, we use five different ways of representing entities, which leads to five different clusterings of them and, thus, five cluster-based types for each entity. By using five different typing predictions together, our system achieves better or comparable results based on randomization tests \cite{Yeh00} with respect to the state-of-the-art levels on four defacto test sets. We publicly share our tools and data sets\footnote{Visit https://tabilab.cmpe.boun.edu.tr/projects/ned\_w\_cmt/ for more information.}.

Here is a summary of our contributions to the literature:
\begin{enumerate}

\item This research is the first at using clustering to obtain cluster-based mention types and using these to improve the NED task.

\item We introduce five different ways of cluster-based mention typing based on representing the context around a mention at three different levels. We show that improved NED results are achieved when the different typing models are used together.

\item In the candidate generation phase, we propose using the candidates of the cooccurring mentions in the same document, which leads to higher gold recall values than previously reported results. 
\end{enumerate}

The rest of this paper is organized as follows. In the next section, we give the related work on NED, mention typing, and clustering. Section 3 introduces cluster-based mention typing, where the methods for clustering named entities and then predicting the cluster-based types are presented. Section 4 describes how to disambiguate the entities. It starts with our candidate generation method and then explains the local context-independent and global context-based features used to represent the mentions for  disambiguation. It ends with the description of our ranking model for disambiguation. Section 5 gives the details on our experimental setup. Next, in Section 6, we present our results on the mention typing model and the disambiguation model, along with a detailed error analysis. Finally, Section 7 concludes the paper with a discussion and future work.

\section{Related Work}

\textbf{Named entity disambiguation} is one of the most studied tasks in the Natural Language Processing (NLP) literature. There are various approaches to formalize the ranking of the entity candidates for disambiguation given an entity mention in text. Early studies \cite{Bunescu06,Mihalcea07} ranked the candidates individually by using  features like similarity between the candidate and the test document. Cucerzan \shortcite{Cucerzan07}, hypothesizing that entities should be correlated and consistent with the topic of the document, used features that optimize the global coherence. Ratinov \textit{et al.} \shortcite{Ratinov11} and Yamada \textit{et al.} \shortcite{Yamada16} used a two-step approach where they select the most likely candidates in the first step, calculate their coherence score based on their surrounding selected candidates and then use that score in the second step (i.e., the ranking step). Due to its simplicity, we also adapt this two-step approach. Later studies \cite{Kulkarni09} considered the ranking collectively, rather than individually. Graph-based approaches \cite{Han11,Hoffart11} were proposed for collective disambiguation, where the topic coherence of an entity is modeled as the importance of its corresponding node in a graph. Several studies were conducted where different algorithms were used to model node importance, including the personalized version of the PageRank algorithm \cite{Pershina15}, probabilistic graphical models \cite{Ganea16}, inter-mention voting \cite{Ferragina10}, random walk \cite{Han11,Guo16}, minimum spanning tree \cite{Phan18}, and gradient tree boosting \cite{Yang18}. Unless heuristics are used, these models are, in general, computationally expensive, as they consider many possible relations between the nodes.

In addition to the way to rank, representing the context, mention and entity is another important aspect in NED. Han and Sun \shortcite{Han12} and Kataria \textit{et al.} \shortcite{Kataria11} used topic modeling to obtain word-entity associations. However, such models learn a representation for each entity as a word distribution. Baroni \textit{et al.} \shortcite{Baroni14} argued that counting-based distributional models are usually outperformed by context-predicting methods, such as embeddings. That said, improved embedding calculations with word2vec \cite{Mikolov13b} led to many studies. Fang \textit{et al.} \shortcite{Fang13}, Zwicklbauer \textit{et al.} \shortcite{Zwicklbauer16a}, and Yamada \textit{et al.} \shortcite{Yamada16} investigated learning entity and word embeddings jointly, which enables a more precise similarity measurement between a context and a candidate. They used the text in Wikipedia as the main source for entity representations. In addition, the knowledge graph of Wikipedia has also been exploited \cite{Radhakrishnan18}. Zwicklbauer \textit{et al.} \shortcite{Zwicklbauer16b} learned multiple semantic embeddings from multiple KBs. In our study, we obtain the entity embeddings with their EAD-KB approach. However, instead of using multiple KBs, we make use of the context of the same KB in five different levels. When it comes to deep learning approaches, Phan \textit{et al.} \shortcite{Phan17} employed Long Short-Term Memory (LSTM) networks with attention mechanism. Ganea and Hofmann \shortcite{Ganea17} used attention mechanism over local context windows to spot important words and Liu \textit{et al.} \shortcite{Liu19} expanded this to the important spans with conditional random fields. While these approaches used neural networks with attention mechanism to model the named entities and mentions together and pick the best matching candidate entity, we used a simple LSTM to model the type prediction of the mentions only and then used that information as an extra clue for a simple feed-forward neural network-based ranking model. Other studies \cite{Sun15,Phan17,Sil18} modeled context using the words located at the left- and right-hand sides of the mention. Either the sentence or a small window is used as the context boundary. Similar to these studies, we model the left and right context separately. In addition, we propose representing the local and global context separately, in three different ways, which in our results is empirically shown to provide a richer way of characterizing entity mentions.

\textbf{Mention typing} is the task of classifying the mentions of named entities with their context dependent types. It is a relatively new study area and a specialized case of corpus-level entity typing, which is the task of inferring all the possible type(s) of a named entity from the aggregation of its mentions in a corpus. Some of the recent studies on corpus-level entity typing used the contextual information \cite{Yaghoobzadeh15}, the entity descriptions in KB \cite{Neelakantan15,Xie16} as well as multi-level representations at word, character and entity level \cite{Yaghoobzadeh17}. The way Yaghoobzadeh and Schutze (2017) represent the entities in terms of these three levels with increasing granularity resembles our way of considering the context at different scales by representing local and global context separately. In case of the mention-level entity typing or mention typing in short, Ling and Weld \shortcite{Ling12} proposed an entity recognizer called FIGER, which uses a fine-grained set of 112 types based on Freebase \cite{Bollacker08} and assigns those types to mentions. They trained a linear-chain conditional random fields model for joint entity recognition and typing. Yosef \textit{et al.} \shortcite{Yosef12} derived a very fine-grained type taxonomy from YAGO \cite{Mahdisoltani15} based on a mapping between Wikipedia categories and WordNet \cite{Miller95} synsets. Their taxonomy contains a large hierarchy of 505 types organized under 5 top level classes (person, location, organization, event, and artifact). They used a support vector machine-based hierarchical classifier to predict the entity mention types. These studies usually created their training data sets from Wikipedia using a distant supervision method, which is the practice that we also employed. Mention typing has also recently been used to improve NED. Raiman and Raiman \shortcite{Raiman18} used Wikipedia categories to incorporate the symbolic information of types into the disambiguation reasoning. Gupta \textit{et al.} \shortcite{Gupta17} used type, description and context representations together to obtain entity embeddings. Murty \textit{et al.} \shortcite{Murty18} employed CNN with position embeddings to obtain a representation of the mention and the context. Onoe and Durrett \shortcite{Onoe19} formulated NED as purely a mention typing problem. However, all of these studies rely on manually crafted type taxonomies. The main difference of our approach from these studies is that we generated types automatically. We use clustering to partition the named entity space of our KB into clusters, each holding entities that occur in a similar context. Then, each cluster is assigned as a type to the entities in that cluster. This makes our cluster-based types more context oriented than manually crafted types. Moreover, since we obtain multiple clusterings based on different contextual scopes, we ended up having multiple type sets, each exhibiting the characteristics of the context differently, unlike the traditional manually crafted type sets in the literature.

\textbf{Clustering} is a powerful tool to partition the data set into similar groups without any supervision. There is a large variety of methods in the literature. They can be mainly grouped into centroid-based clustering, such as K-means \cite{Steinley06}, density-based clustering, like DBSCAN \cite{Ester96}, distribution-based clustering, like the Expectation-Maximization algorithm \cite{Jin11}, and hierarchical clustering. Among them, centroid-based clustering, and more specifically K-means, is one of the most practical algorithms due to its simplicity and time complexity. One of the early application areas of clustering includes the clustering of search results \cite{Van79} in the  Information Retrieval field. Later studies categorized named entities in order to improve document retrieval \cite{Pasca04,Teffera10}. In the NLP field, clustering has been used to group similar words together. Brown and Mercer \shortcite{Brown92} introduced Brown clustering which assumes that similar words have similar distributions of words to their immediate left and right. While this method assigns each word to one cluster (i.e., hard clustering), Pereira \textit{et al.} \shortcite{Pereira93} proposed a distributional clustering method which calculates the probability of assigning a word to each cluster (i.e., soft clustering). Word clustering has been used to improve many tasks like statistical language modeling \cite{Kneser93}, text classification \cite{Slonim:01}, and induction of part-of-speech tags \cite{Clark03}. In the Named Entity Recognition task, Ren \textit{et al.} \shortcite{ren15} clustered text patterns that represent relations between certain types of named entities in order to better recognize their mentions in text. In mention typing, Huang \textit{et al.} \shortcite{Huang16} proposed a joint hierarchical clustering and linking algorithm to cluster mentions and set their types according to the context. Their approach to clustering is similar to ours. However, they rely more on knowledge base taxonomy in order to generate human-readable types. In our case, we do not need to have human-readable types, as mention typing is merely an intermediate step in order to obtain additional contextual clues for  named entity disambiguation. In the NED task, the term ``Entity Clustering'' has exclusively been used for the co-reference resolution (CR) task \cite{Cardie99}, which is to detect and group the multiple mentions of the same entity within a document or multiple documents (\textit{i.e.} cross-documents). If this task is done on mentions that have no corresponding entries in KB, it is called ``NIL clustering'' \cite{Ratford11}. In these studies, hierarchical agglomerative clustering is mainly used due to its efficiency as it merges similar mentions into a new group recursively in a bottom-up approach. When CR is done within a document, the clustering only considers the merge combinations within that document, which can be in the order of thousands. However, the number of combinations in cross-document CR can be in the order of millions, which requires more efficient clustering algorithms. Some of the proposed methods include a distributed Markov-Chain Monte Carlo approach to utilize parallel processing \cite{Singh11}, a discriminative hierarchical model that defines an entity as a summary of its children nodes \cite{Wick12} and the use of latent features derived from matrix factorization of the surrounding context \cite{Ngomo14}. Moreover, Singh \textit{et al.} \shortcite{Singh10} proposed a discriminative model which is trained on a distantly-labeled data set generated from Wikipedia. A recent review of the CR literature is provided by Beheshti \textit{et al.} \shortcite{Beheshti17}. CR has also been used in a joint task with entity linking \cite{Monahan11,Dutta15}. Apart from using mention clustering  directly, Hendrickx and Daelemans \shortcite{Hendrickx07} clustered 10000 lemmatized nouns into 1000 groups based on syntactic relations in order to learn features that are useful for the CR task. While clustering has been explicitly used on mentions of named entities, to the best of our knowledge, our work is the first study on clustering millions of named entities. Moreover, we represent entities at different contextual levels and do the clustering for each level.

\section{Cluster-based Mention Typing}
\label{chap:clustering}

In the mention typing task, named entities in a KB are in general assumed to be assigned to manually-crafted predefined type(s), and the task is to classify the mentions of named entities to these predefined types based on their context. Considering that there can be millions of named entities in a KB, manually designed predefined types are inherently incomplete. Therefore, in this study we propose using clustering to define the entity types in an unsupervised fashion. We cluster named entities that occur in similar contexts together and assign the corresponding cluster id of an entity as a type label to that entity. As Mikolov \textit{et al.} \shortcite{Mikolov13b} argued, similar words occur in similar contexts, hence have similar word embeddings. That said, similar entities should have similar entity embeddings. When we cluster named entities based on these embeddings, similar entities are expected to be grouped into the same cluster and each cluster is expected to contain entities of similar types based on common contextual clues. Note that these cluster-based types do not necessarily correspond to regular named entity types, and they do not need to. In our work, their only purpose is to represent the context so that the named entity disambiguation model can decide how likely it is that a certain candidate named entity is mentioned in the given context.

\begin{figure}[h]
\centering
\includegraphics[width=90mm]{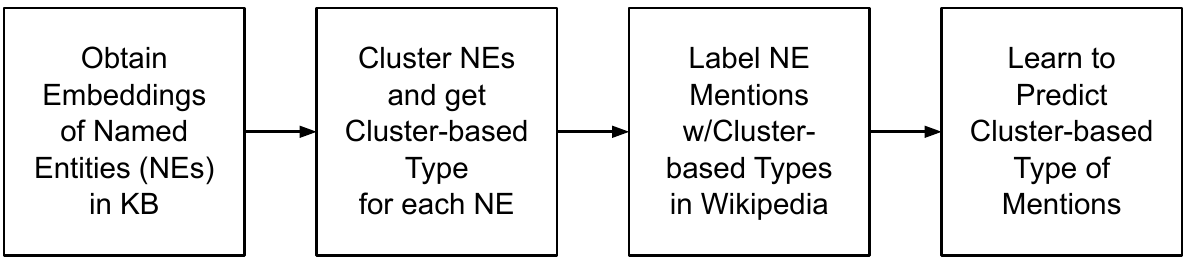}
\caption{The Four Steps of  Cluster-based Mention Typing.}
\label{fig:ClusterBasedMentionTyping}
\end{figure}

As depicted in Figure~\ref{fig:ClusterBasedMentionTyping}, our cluster-based mention typing involves four steps. First, we calculate the entity embeddings based on some contextual representation. In the second step, clustering is applied to group similar entities based on their embeddings. As we get the clusters, we assign the cluster id as a cluster-based type to each entity in that cluster. In the third step, to train a typing model, we prepare a training data set by automatically labeling the hyperlinked mentions of named entities in Wikipedia articles with the assigned cluster-based types. In the final step, we train a typing model with this auto-generated training data. At test time, the typing model only uses the mention surface form and its surrounding context to predict the cluster-based type. These cluster-based type predictions are used as extra features at the entity disambiguation step.

In this study, we cluster entities in five different ways based on different representations of entities. We obtain four embeddings for each entity based on four different (three contextual, one synset based) representations and use them to produce four separate clusterings of the entity space with K-means. To increase the variety, we also use Brown clustering, which requires no embeddings but takes simpler input. At the end, we get five different mention typing models, one for each clustering. Before explaining each step in the following subsections, we first describe how to represent the context of a mention in three different formats.

\subsection{Three Different Representations of a Mention's Context}
\label{Sec:MentionContext}

\begin{figure}[h]
\centering
\includegraphics[width=105mm]{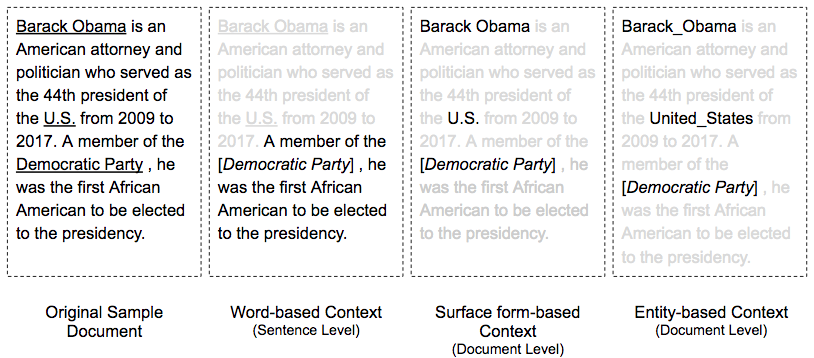}
\caption{A Sample Document and the Different Types of Context used for the Named Entity Mention of ``Democratic Party".}
\label{fig:Context}
\end{figure}

The context of a mention is basically the content that resides at the left- and right-hand side of the mention. These are called left and right contexts, respectively. In this research, we represent this context in three different ways. Figure~\ref{fig:Context} shows a document at the left-hand side where three entity mentions are underlined. At its right, it shows the three different ways of representing the context specific to the mention ``Democratic Party". For that particular mention, the dimmed sections are not part of the used context.

\textbf{Word-based Context (WC)} is a traditional context in the form of a sequence of words adjacent to the mention at its left- and right-hand side. This context has a local viewpoint, since only the words of the sentence that holds the mention are used. This is shown in the second box of Figure~\ref{fig:Context}, where the words in the same sentence with the mention are kept as the word-based context. \textbf{Surface Form-based Context (SFC)} keeps the surface forms of the mentions at the left- and right-hand side of each mention, excluding all other words in between as shown in the third box. Compared to WC, this includes words farther than the mention into the context. They reflect the topic of the document better than the other words.  \textbf{Entity-based Context (EC)} only consists of entity ids surrounding the mention, excluding all the words as shown in the fourth box. Considering the fact that cooccurring entities are related and consistent with the topic of the document, EC also reflects the topic of the document. Both SFC and EC present a global viewpoint at the document level compared to the local viewpoint of WC that is mostly based on the sentence level. Having said that, cluster-based types generated with WC-type context may act more like traditional named entity types, as the surrounding words might reflect their semantic roles better. On the other hand, cluster-based types based on SFC and EC may act more like topic labels.

\subsection{Obtaining Entity Embeddings}
\label{Sec:EntityEmbeddings}

\begin{figure}[h]
\centering
\includegraphics[width=85mm]{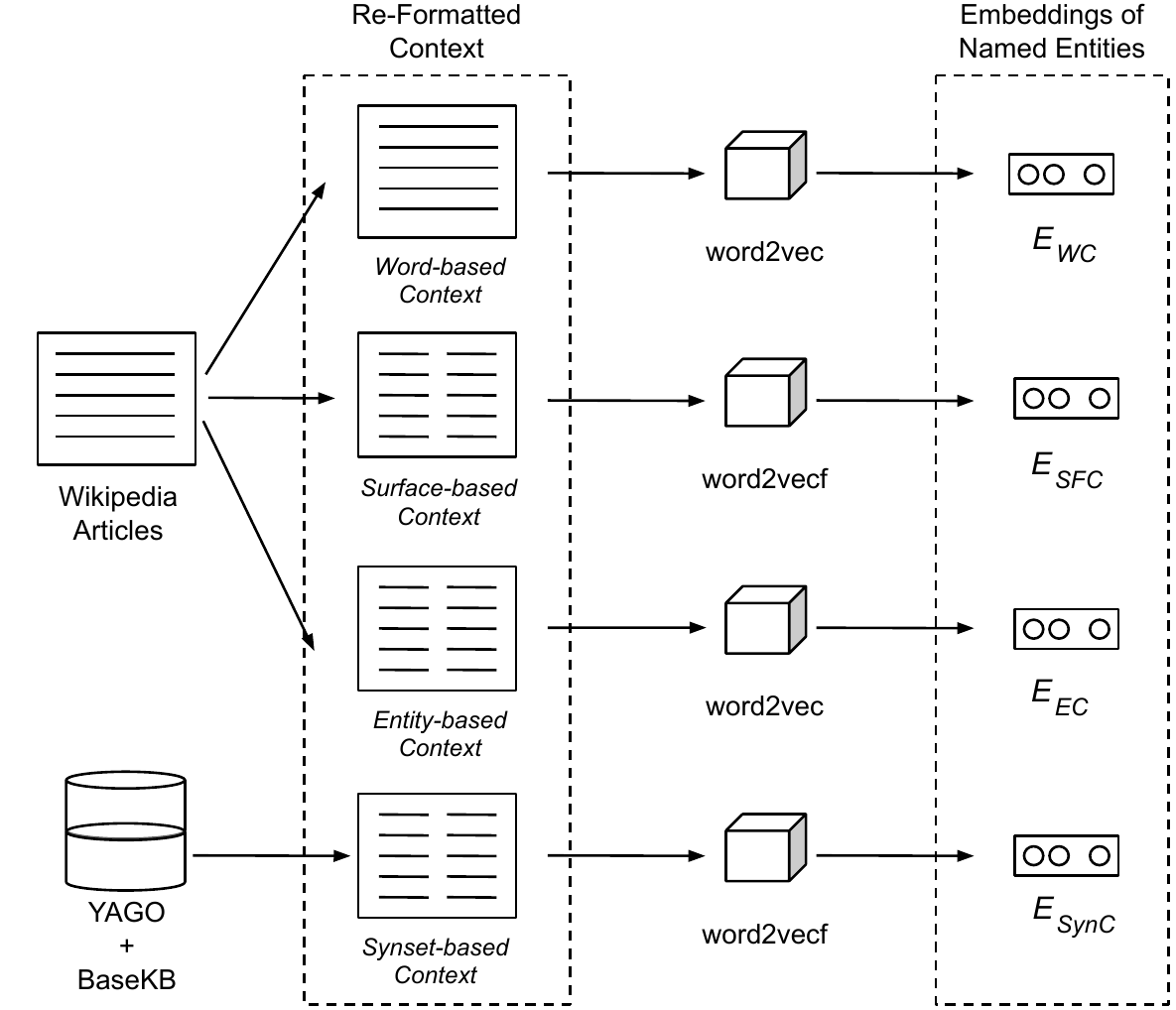}
\caption{Steps of Obtaining Entity Embeddings by Using Different Types of Context.}
\label{fig:ObtainingEntityEmbeddings}
\end{figure}

We obtain four different embeddings for each entity in KB by using the Wikipedia articles as training data. Three of those embeddings are based on the three different ways of representing the context of the named entity mentions in text. The fourth embedding is based on the synsets (or types in BaseKB terminology) in the YAGO and BaseKB data sets that are associated with each entity. Figure~\ref{fig:ObtainingEntityEmbeddings} overviews the process.

\textbf{WC-based Entity Embeddings} ($E_{WC}$) are obtained by using the \textit{word2vec} tool \cite{Mikolov13b} on the Wikipedia articles. \textit{Word2vec} gets the input as a sequence of tokens and calculates an embedding for each token (i.e., target token) in the given input based on its window of surrounding tokens at the left- and right-hand sides. Note that \textit{word2vec} does not use the sentence boundaries as context boundaries as in the definition of WC. Instead, it uses a window of words around the mention. In order to obtain the embeddings for the entities, we reformat the Wikipedia articles based on the EAD-KB approach \cite{Zwicklbauer16b}. We replace the hyperlinked mention (i.e., a-href tag and the surface form together) of each named entity with its Wikipedia id, which is a single unique token that corresponds to that entity. As we run the \textit{word2vec} tool on these reformatted articles, we obtain embeddings for both regular words and entity ids. 

\begin{figure}[h]
\centering
\includegraphics[width=95mm]{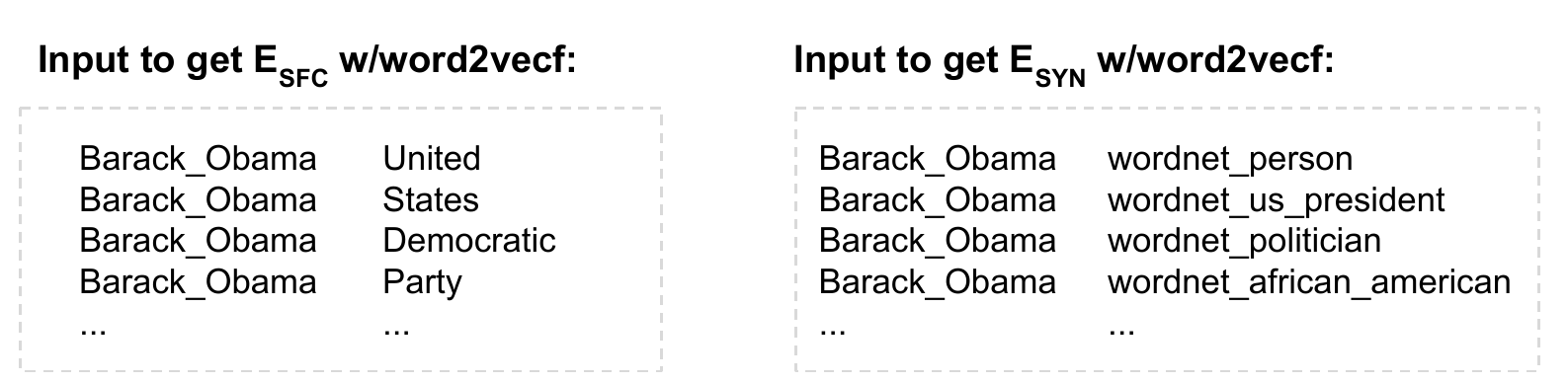}
\caption{Input Samples from the Example Contexts in Figure 3 to Obtain $E_{SFC}$ and $E_{SYNC}$ with the word2vecf Tool.}
\label{fig:Context2}
\end{figure}

\textbf{SFC-based Entity Embeddings} ($E_{SFC}$) do not rely on the immediate adjacent words of the entity mention as in $E_{WC}$. It is hard to represent this in a linear bag-of-words context as \textit{word2vec} expects. Hence, we use the \textit{word2vecf} tool \cite{Levy14}, which is an altered version of \textit{word2vec}. While \textit{word2vec} assumes a window of tokens around the target token as context, \textit{word2vecf} allows us to define the context tokens arbitrarily, one-by-one for the target token. It takes the input in two columns, where the first column holds the target token and the second column has a single context token. As exemplified in Figure~\ref{fig:Context2}, the input file contains one row for each target and context token pair. We select a window of mentions around each hyperlinked mention in Wikipedia and use the words in their surface forms to create the input data for the \textit{word2vecf} tool.

\textbf{EC-based Entity Embeddings} ($E_{EC}$) are calculated by using the entity ids of the surrounding mentions in the given EC. In this case, we remove all the words in the Wikipedia articles and only keep the entity ids of hyperlinked mentions as tokens. As we run \textit{word2vec} on these reformatted articles, we get an embedding for each mentioned named entity, which is calculated based on a window of the mentioned named entities around it.

\textbf{Synset-based Entity Embeddings} ($E_{SYN}$) are the only type of entity embeddings that do not reflect the context-based similarity of entities, but their synset-based similarity. In WordNet, a synset is a set of synonymous words grouped together. WordNet uses the synset records to define a hypernym hierarchy between them to reflect which synset is a type of which other. YAGO v3.1 \cite{Mahdisoltani15} uses those synsets as category labels to represent certain types of named entities, such as person, musician, city etc. Moreover, YAGO extends the WordNet synsets with its own specialized synsets, prefixed ``wikicat'' (e.g., wordnet\_person\_100007846 is a hypernym of wikicat\_Men in the YAGO synset taxonomy). In YAGO data set, named entities are labeled with one or more synsets. In addition to YAGO, we also use BaseKB Gold Ultimate Edition \cite{Bollacker08} data set which is based on the last Freebase dump of April 19, 2015. It contains over 1.3 billion facts about 40 million subjects, including 4 million from Wikipedia. It is similar to YAGO, except it has its own simple type taxonomy, independent of WordNet synsets. In our experiments, we combine the type definitions from both YAGO and BaseKB data sets and call them synsets for the sake of simplicity. By combining them, we aim to have a synset for as many named entities as possible. We then use the associated synsets of named entities as their context tokens, as \textit{word2vecf} allows us to define custom context. We give the entity ids in the first column and the associated types in the second column as shown in Figure~\ref{fig:Context2}. In this process, we do not use all available synsets though. We replace any specialized synset (ones starting with wikicat\_*) with its hypernym WordNet synset. We also filter out some BaseKB types that do not reflect type information (e.g., base.tagit.topic and types with user names).

\subsection{Clustering Named Entities}
\label{Sec:EntityClustering}

Now that we have entity embeddings, we can use them to cluster named entities. In order to do that, we primarily use the K-means algorithm due to its simplicity and time complexity. It is a centroid-based clustering algorithm. After setting the number of centroids (i.e. $K$) manually at the beginning, it assigns every data point to the nearest centroid. In this study, we use the euclidean distance between the entity embedding and the centroid vectors. After the assignment step, the centroids are recomputed. This process continues till a stopping criterion is met\footnote{We stop iterating when there is no more than 1\% change in the assignment of entities to the clusters. We also allow at most 50 iterations.}. In addition to K-means, we use the Brown clustering which is originally introduced to group words that have similar distributions of words to their immediate left and right. Both algorithms have time complexity that is linear in terms of the number of items being clustered \cite{Steinbach00} provided that other factors are fixed. Considering that we have over five million named entities in KB, this property makes them very eligible for our experiments.

\begin{figure}[h]
\centering
\includegraphics[width=60mm]{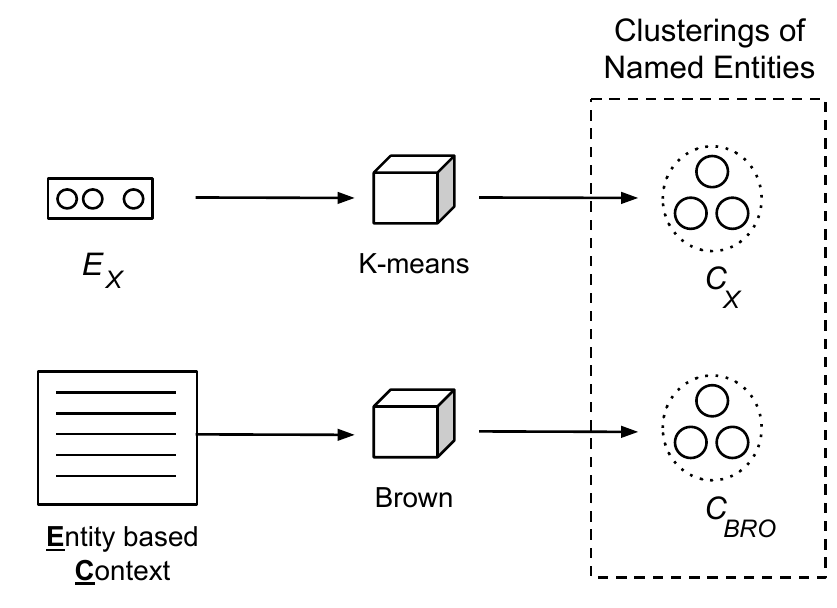}
\caption{Creating Different Clusterings of Named Entities with the K-means and Brown Clustering Algorithms, where $X\in$\{WC,SFC,EC,SynC\}.}
\label{fig:EntityClustering}
\end{figure}

As in Figure~\ref{fig:EntityClustering}, we give entity embeddings ($E_X$) to K-means as input. In case of Brown, we use the Wikipedia articles in the Entity-based context format. After getting clusters ($C_X$), the cluster ids are assigned to entities as cluster-based types.

It is important to note that each clustering of the named entity space breaks that space into groups. In terms of the named entity normalization task, the discriminative power of a clustering depends on how well it distinguishes the correct candidate for a named entity mention from the other candidates. The ideal case occurs when the correct named entity for a mention is placed in a different cluster from the other most likely candidates and the cluster of the mention is also predicted to be the same as its corresponding named entity. Using a high number of clusters makes the clustering more discriminative as each cluster corresponds to a lower number of entities. However, that also makes the typing model task harder as it increases the ambiguity. In Section~\ref{Sec:ClusterOptimization}, we explore the right number of clusters for our experiments. Moreover, since we have five different clusterings and each breaks the entity space differently, using a combination of them is expected to make the aggregate discriminative power even higher. Hence, using multiple clusterings can be seen as an alternative to using a higher number of clusters. Note that Section~\ref{Sec:ClusterOptimization} also explains our heuristic to select the best combination (in terms of number of clusters for each clustering) in order to achieve a better performance at the disambiguation step.

\subsection{Preparing Training Data for the Typing Model}
\label{Sec:PreparingTrainingData}

So far, we have described our approach for clustering named entities and assigning the cluster ids to entities as types. In order to train a typing model to predict the type of an entity mention in an input text, we need training data that includes mentions labeled with these types. We create the training data from the hyperlinked mentions in Wikipedia articles using distant supervision. We label each hyperlinked named entity mention with its assigned cluster-based type. Since we have five different clusterings, we produce five different training data sets. 

\begin{figure}[h]
\centering
\includegraphics[width=125mm]{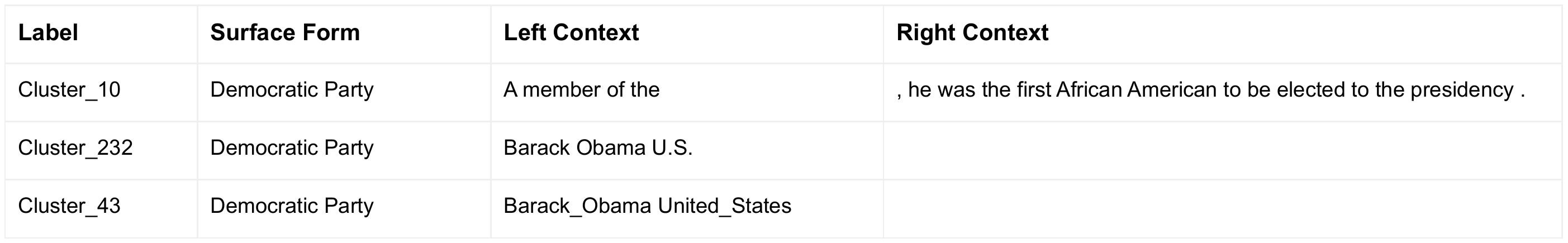}
\caption{Example Training Instances from Figure 3 in WC, SFC, and EC Formats, Respectively.}
\label{fig:ExmapleTypingModelTrainingData}
\end{figure}

As described in Section~\ref{Sec:MentionContext}, we represent the context of a mention in three different formats and three of our clusterings (namely, $C_{WC}$, $C_{SFC}$, $C_{EC}$) are obtained according to the corresponding context formats. Take $C_{WC}$ for example. Clusters in $C_{WC}$ hold entities that are similar to each other in terms of word-based context (WC). It is compatible to train a typing model for $C_{WC}$ with input in the WC format because labels are created from the characteristics that made up the input in that format. Same is true for $C_{SFC}$ that goes with the surface form-based context (SFC) and $C_{EC}$ with entity-based context (EC). Figure~\ref{fig:ExmapleTypingModelTrainingData} exemplifies three training instances in WC, SFC, and EC formats, respectively. They are generated based on the example in Figure~\ref{fig:Context}. Each instance consists of three input fields in addition to the label; surface form of the hyperlinked mention, as well as left and right context of that mention. Surface form is the common input. Like surface form, contexts in the first two cases are in the form of words as exemplified in Figure~\ref{fig:ExmapleTypingModelTrainingData}. In case of the third case, it is in entity ids.

In case of $C_{SynC}$ and $C_{BRO}$, we use word-based context and surface form-based context, respectively. In order words, we use the same formatted input for $C_{WC}$ and $C_{SynC}$ pair and $C_{SFC}$ and $C_{BRO}$ pair with the exception of the labels. Note that clusters in the $C_{SynC}$ hold entities that are associated with similar synsets. Synsets are like semantic categories and local context is better suited to infer which synsets the mentioned entity is associated with. Hence, word-based context is used. On the other hand, using global context is better suited to infer $C_{BRO}$ based types. We choose to use the surface form-based context over the entity-based context. This is due to the fact that surface form-based context can easily be obtained by gathering the surface forms of the surrounding named entities. However, entity-based context is only available after we get the best predictions from the first stage of our two-stage disambiguation approach described in Section~\ref{Sec:DisambiguatingNamedEntities}. In other words, by using the surface form-based context, we are making the Brown-based mention typing model available at the first stage, which increases the success rate of the first stage.

\subsection{Specialized Word Embeddings for the Typing Model}
\label{Sec:SpecializedWordEmbeddings}

When it comes to predicting the cluster of a mention with word-based context, it is common practice to use word embeddings that are obtained from a data set that is similar to the domain data. In our case, it is Wikipedia, which is a widely used source to obtain word embeddings in the literature. In our experiments, we did not use regular word embeddings calculated with \textit{word2vec}. Instead, we propose two word embeddings that are more discriminative at predicting the cluster of a given mention. We use these word embeddings together to represent the word-based input for the mention typing model.

\textbf{Cluster-centric Word Embeddings} ($W_{CC}$): We hypothesize that word embeddings, which are obtained from the data that reflect the characteristics of the problem, may be more effective at solving that problem. In this case, the problem is to predict the cluster-based type of a mention. Hence, we inject a piece of cluster information into the context in hope that word embeddings are being influenced by their presence and become better at predicting the cluster-based type. In order to accomplish that, in Wikipedia articles, we filter out the html tag (i.e., a-href) of the named entity mentions and leave its surface form alone. At the same time, we add a special token to the right-hand side of that surface form. That token corresponds to the assigned cluster id of the mentioned named entity. Embeddings of those words that are close to the specific cluster token are expected to be affected by that and become indicators of that cluster. We use the \textit{word2vec} tool on this modified data set. Since the clustering of named entities is done at training time and does not change based on the given test input, the same embeddings that are obtained at training time are used at the test time. We obtain different word embeddings for each of the five named entity clusterings, since entities are assigned to different clusters in each case.

\textbf{Surface Form-based Word Embeddings} ($W_{SF}$): While word embeddings in $W_{CC}$ are only based on the words surrounding the named entity mention, another type of word embedding can be obtained with the words inside the surface forms exclusively. In order to do that, we use the Anchor Text data set\footnote{We use the English version of the Anchor Texts data set from  DBPedia Downloads 2016-10 available at https://wiki.dbpedia.org/downloads-2016-10.} of DBpedia \cite{Auer07}. It contains over 150 million (entity\_id, surface\_form) records, one for each mention occurrence in Wikipedia. We extract around fifteen million unique such records for almost all named entities in our KB, along with their frequencies. We name this data set as the \textbf{Surface Form Data Set}. Since this data set only includes the surface forms of the entities, but not their surrounding sentences, we use the associated cluster ids of entities as context tokens with the word2vecf tool. In other words, for each word in the surface form of a named entity, the cluster id of that entity is given as a context word, which results in  (surface\_form\_word, cluster\_id) pairs. We also take the frequencies of the surface form records into account. The more frequent one surface form is, the more important it is. Therefore, while creating the training data set for $W_{SF}$ from the (surface\_form\_word, cluster\_id) pairs, the number of instances included for each surface\_form\_word is proportional to the logarithm of its frequency in the Surface Form Data Set.

\subsection{Model to Predict Cluster-based Types}
\label{Sec:ClusterPredictionModel}

Now that we generate training data that includes labeled mentions with their corresponding named entities' cluster-based type, we are ready to train a mention typing model to predict those types. At the end, we end up having five typing models: \textit{Word} model based on $C_{WC}$, \textit{Surface} model on $C_{SFC}$, \textit{Entity} model on $C_{EC}$, \textit{Brown} model on $C_{BRO}$ and \textit{Entity} model on $C_{EC}$.

We experiment with two different typing model architectures: Long Short-Term Memory (LSTM) and Convolutional Neural Network (CNN) models. CNNs are widely used for text classification, whereas LSTMs are effective at modeling sequences of items. Both models receive the same three-part input described in the previous section. However, since we have two sets of word embeddings for surface words as described in \ref{Sec:SpecializedWordEmbeddings}, we can represent the Surface-part as two inputs for both models, one with $W_{CC}$ and the other with $W_{SF}$. In our experiments, we seek to compare both models and pick the best performing one. Details on these models are given below.

\subsubsection{LSTM-based Typing Model}
\begin{figure}[h]
\centering
\includegraphics[width=105mm]{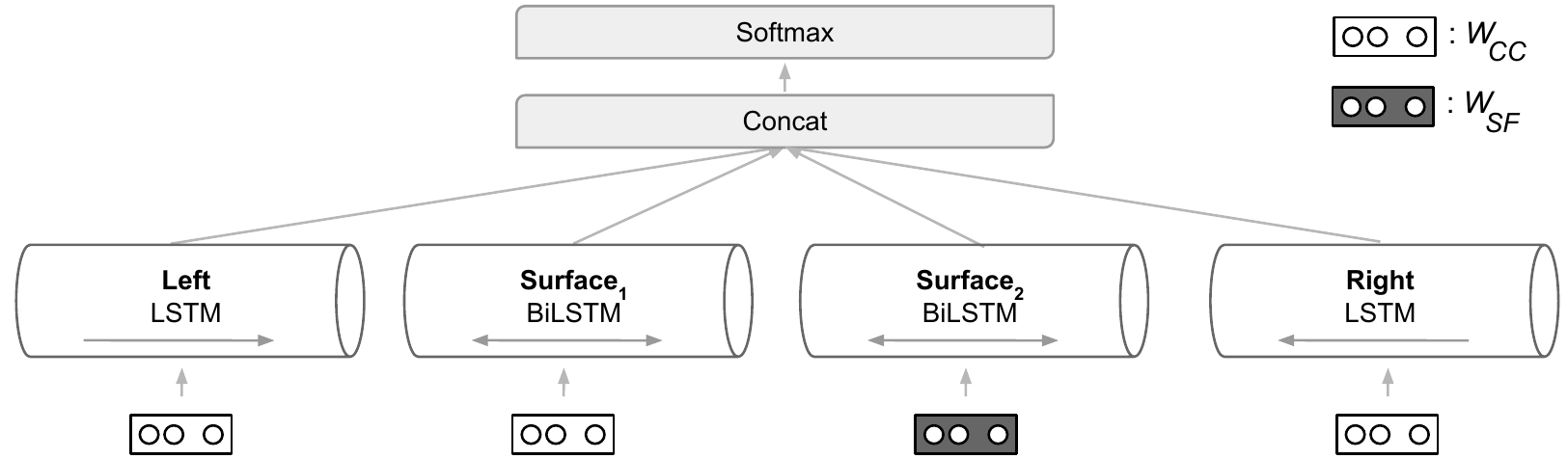}
\caption{LSTM-based Model for Mention Typing.}
\label{fig:LSTMPredictionModel}
\end{figure}

In this model, each input part is given to a separate LSTM layer and their output is concatenated and fed into the final softmax layer as illustrated in Figure~\ref{fig:LSTMPredictionModel}. $Surface_1$ gets the Cluster-centric Word Embeddings ($W_{CC}$), which are better at discriminating cluster-specific words. $Surface_2$ gets the Surface Form Word Embeddings ($W_{SF}$), which are optimized for discriminating cluster-specific surface words seen in large set of surface forms. We use dropout layers \cite{Srivastava14} before and after the LSTM layers in order to prevent overfitting. Our preliminary experiments led us to using bi-directional LSTM (BiLSTM) for the $Surface_x$ and uni-directional one for the left and right ones. Moreover, reversing the order of words in the right context produces better results. We argue that words that are closer to the mention are more related to the mention and play a more important role at the prediction.

\subsubsection{CNN-based Typing Model}
\label{Sec:CNNbasedTypingModel}
\begin{figure}[h]
\centering
\includegraphics[width=105mm]{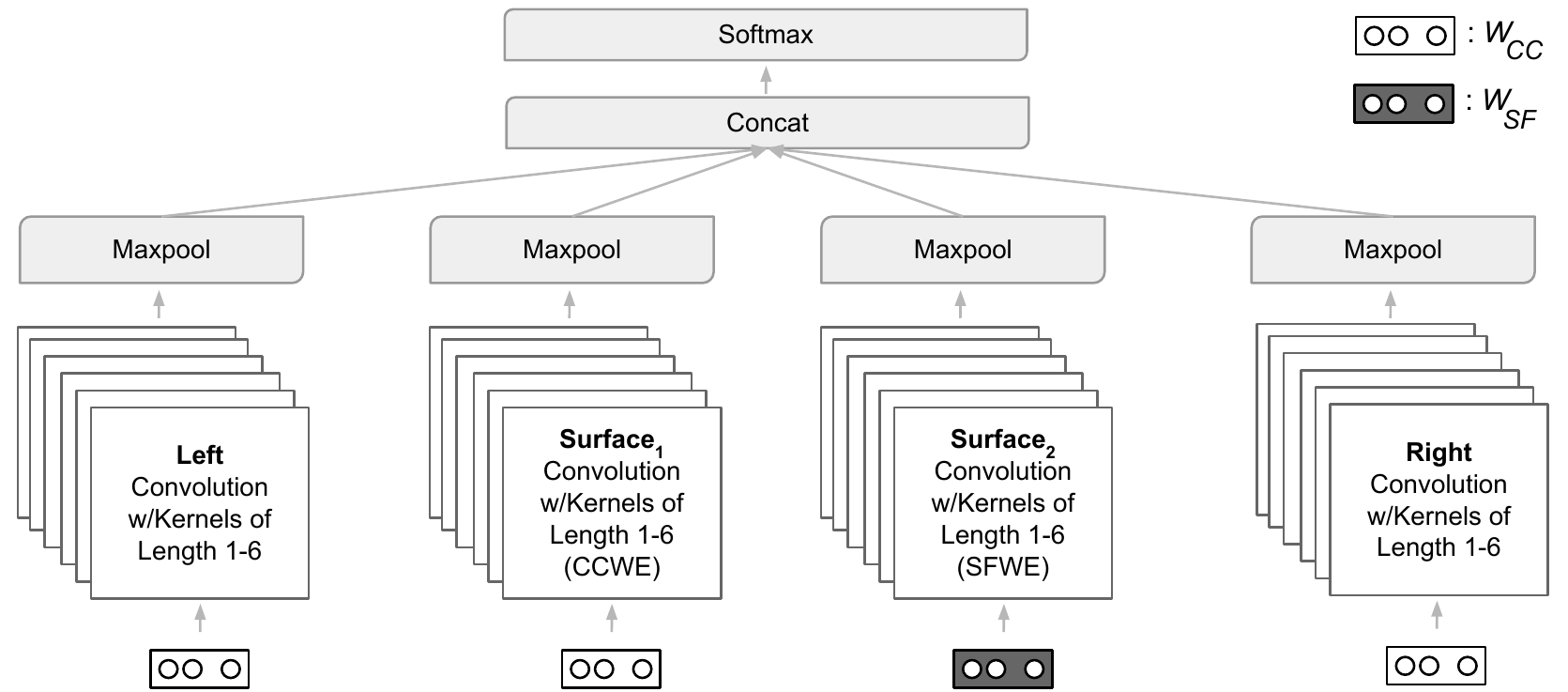}
\caption{CNN-based Model for Mention Typing.}
\label{fig:CNNPredictionModel}
\end{figure}

CNN-based model is very similar to the LSTM-based model, except that the input parts are given to the convolution layers. As pictured in Figure~\ref{fig:CNNPredictionModel}, each input part is fed into the six convolution layers with kernel sizes 1 through 6. These layers model the sequence of words in different lengths, like n-grams in a language model. We use hyperbolic tangent as an activation function over the convolution layers and apply max pooling layer prior to merging all into one layer before feeding to the softmax layer. Like in the LSTM-based model, we use dropout after the word embeddings and before the softmax layer.

\section{Disambiguating Named Entities}
\label{Sec:DisambiguatingNamedEntities}

Disambiguation of named entities is often formulated as a ranking problem. First, a set of candidate named entities for a given mention are obtained. Then, each of them is scored individually, resulting in a ranked list of candidates. Finally, the highest ranked candidate is returned as the final prediction. In our experiments, we use a two-stage approach for ranking. At the first stage, we train our model and rank the candidates with the available features and get a ranking probability for each candidate. As a result of the first stage, the highest scored candidates are used to define the entity-based context shown in Figure~\ref{fig:Context}. This allows us to run the \textit{Entity} typing model, which accepts the context in terms of entities only. At the second stage, we use those ranking probabilities to define new features that encapsulate  future insight. With the addition of new features, we again train our ranking model and get the final ranking of the candidates. In the following subsections, we first describe how we obtain the candidates. Next, we explain our disambiguation model by first describing the features we used and then, the model itself.

\subsection{Candidate Generation}
\label{Sec:CandidateGenerator}
In the literature, most of the studies do not mention how to index and search entities to generate candidates based on the surface form of a mention. Some of the ones \cite{Hachey13,Hakimov16} providing this information report that they use Apache Lucene\footnote{Available at http://lucene.apache.org/}. Lucene provides fuzzy search property which tolerates a certain percentage of mismatch at character level with the Damerau-Levenshtein distance. However, such off-the-shelf tools consider each mention individually and do not use contextual clues from other mentions in the same document. Moreover, they do not provide additional information, such as the type of the matched surface form, which we use as a feature (i.e. \textit{surface-form-type-in-binary} in Table 1) at the disambiguation step. Hence, we implemented our own candidate generation tool. 

For the indexing, as in \cite{Hoffart11}, we use all available surface forms of named entities in DBpedia and Wikipedia (including anchor texts, redirect and disambiguation records). In other studies, while some \cite{Ling15,Ganea17} used additional web corpora, others \cite{Ratinov11,Phan18} used only Wikipedia. We index surface forms based on character tri-grams. In case of searching, the pseudocode of our search procedure is in Algorithm 1.

Our algorithm works in two stages. In the first stage, it starts with searching named entities matching with the mention surface form (Lines 17-25). It first retrieves indexed surface forms that have at least a certain amount of character tri-gram overlap (T=60\%) with the search query (i.e., mention surface form). Then, it picks the ones that have the ratio of the edit distance at the character level to the query length less than certain value (E=25\%). If not, then it checks for the word overlap. It can tolerate up to one word mismatch (D=1) if the query contains more than one word (W=2). After this selection, all matched candidates are returned. This first stage is similar to what off-the-shelf third party indexing and search tools provide.

\begin{algorithm}
\small
\caption{Pseudocode of the candidate generation algorithm}\label{euclid}
\begin{algorithmic}[1]

\Procedure{getMatchedNamedEntitiesForDocument(set\_of\_mentions, N)}{}
\State \textit{matchedEntitiesForMentions} = []
\For{\textbf{each} mention \textbf{in} set\_of\_mentions}
\State matchedEntities = getCandidatesForMention(mention)
\State add \textit{matchedEntities} to \textit{matchedEntitiesForMentions}
\EndFor
\For{\textbf{each} mention\textsubscript{a} and mention\textsubscript{b} pair \textbf{in} set\_of\_mentions}
\If{mention\textsubscript{a} is completely seen in mention\textsubscript{b}}
\State add \textit{matchedEntities\textsubscript{b}} to \textit{matchedEntities\textsubscript{a}}
\EndIf
\EndFor
\For{\textbf{each} mention\textsubscript{a} in set\_of\_mentions}
\For{\textbf{each} matched\_entity in other mentions}
\For{\textbf{each} most frequently cooccurring entity \textbf{of} matched\_entity}
\For{\textbf{each} surface\_form \textbf{of} cooccurring entity}
\If{mention\textsubscript{a} is completely seen in surface\_form}
\State add that \textit{entity} to \textit{matchedEntities\textsubscript{a}}
\EndIf
\EndFor
\EndFor
\EndFor
\EndFor
\State \textbf{return} scoreCandidatesThenPickTopN(\textit{matchedEntitiesForMentions}, N)
\EndProcedure

\Procedure{getCandidatesForMention(query)}{}
\State \textit{matchedEntities} = []
\For{\textbf{each} \textit{entity} \textbf{in} the knowledge\_base}
\For{\textbf{each} \textit{surface\_form} \textbf{of} \textit{entity}}
\If{ trigramOverlap(\textit{surface\_form}, \textit{query}) $\geq$ $T$\%}
\If{editDistance(\textit{surface\_form}, \textit{query}) $\le$ $E$\% of \textit{query} length}
\State add \textit{entity} to \textit{matchedEntities}
\ElsIf{ numWords(\textit{surface\_form}) $\geq$ $W$ and numWordDiff(surface\_form, query) $\le$ $D$}
\State add \textit{entity} to \textit{matchedEntities}
\EndIf
\EndIf
\EndFor
\EndFor
\State \textbf{return} \textit{matchedEntities}
\EndProcedure

\end{algorithmic}
\end{algorithm}

In the second stage of our algorithm, we expand the candidates of each mention with the candidates from other mentions in the same document, since the same entity can be mentioned more than once. We do this in two steps. The first step looks at if the surface form of one mention is completely seen in the other's (Lines 6-8). For example, ``Barack Obama'' can be first mentioned in full name and then as ``Obama'' in the same document. The second expansion step uses the most frequently cooccurring named entities (Lines 9-14) observed in Wikipedia articles. We only include the ones that have a surface form that includes the mention surface form in itself. At the end, we order the final set of candidates by scoring each candidate based on a specific formula\footnote{We observed that the following candidate scoring gave the highest recall on the AIDA.train set: score = \textit{entity\_frequency} + \textit{num\_of\_occ\_in\_test\_doc}*100 - \textit{jaro\_winkler\_distance}*10000, where \textit{entity\_frequency} is the total number of times the entity is seen in our Surface Form Data Set. We observe that Jaro-winkler distance performs better than Levenstein edit distance in this final scoring stage.} (Line 15) and return the highest scored 100 candidates as the final output.

\subsection{Ranking Features for Disambiguation}
\label{Sec:RankingFeatures}

Each candidate is scored according to certain properties. These properties are named as features and their scalar values are turned into a vector which is given to the ranking model as input. In our experiments, we use a total of twelve features. In order to understand how they are calculated, we first give the definitions of the relevant variables, sets of values, and functions in Table~\ref{RankingFeatureVariables}.

\begin{table}[h]
\begin{center}
\scriptsize
\renewcommand\arraystretch{1.1}
\begin{tabular}{ll} \hline \hline
\bf Variable: & \bf Description \\ 
$SF_m$ & Surface form of the mention \\
$SF_c^{best}$ & Closest surface form of the entity $c$ wrt $SF_m$ given by the candidate generator\\
$D_c$ & Doc2vec embedding of the candidate entity $c$ \\
$D_t$ & Doc2vec embedding of the test document $t$ \\
$E_c$ & Entity embedding of the candidate entity $c$ \\
$T_c^{SF}$ & Type of the surface form $SF$ given by the candidate generator\\
$T_c^{e}$ & Type of the candidate entity $c$ \\
$P_c^n$ & Probability given by the typing model $n$ for the entity $c$ being in its cluster-based type \\
$R_c$ & Ranking probability of candidate entity $c$ calculated by the first stage ranking \\ \hline
\bf Set: & \bf Description \\ 
$S_c$ & Set of all surface forms for the candidate entity $c$ seen in Surface Form Data Set \\
$C^m$ & Set of all candidates for the mention $m$\\
$C_c^{t}$ & Set of all occurrences of the same candidate $c$ in test document $t$ \\
$C_c^{prev}$ & Set of all previous occurrences of the candidate $c$ wrt the mention $m$ \\
$C_c^{next}$ & Set of all future occurrences of the candidate $c$ wrt the mention $m$\\
$C_{topNxM}$ & Set of highest ranked N candidates in the surrounding window of  M mentions \\  \hline
\bf Function: & \bf Description \\ 
$freq(SF_c)$ & Number of times the surface form $SF$ is seen for entity $c$ in Surface Form Data Set \\
$dist(A, B)$ & Levenstein edit distance between two strings $A$ and $B$ \\
$cos(A, B)$ & Cosine similarity between two vectors $A$ and $B$ \\
$log(A)$ & Natural logarithm of the scalar value $A$ \\
$avg(\sum)$ & Average of the sum of the values \\
$binarize(A)$ & Binarized code of the number $A$ \\
$argmax(S)$ & The maximum value among the set of values $S$ \\
$argsecmax(S)$ & The second maximum value among the set of values $S$ \\
 \hline \hline
\end{tabular}
\end{center}
\caption{\label{RankingFeatureVariables} Variables, Sets, and Functions used at Defining Ranking Features.}
\end{table}
 
\textbf{Variables} in Table~\ref{RankingFeatureVariables} represent either strings, scalar values, or vectors. $SF_m$ and $D_t$ are the only variables that are not related to the candidate. Whereas, $D_c$, $E_c$, and $T_c^e$ are candidate-specific and calculated offline, hence they do not depend on the mention being considered. The rest of the variables depend on the mention. \textbf{Sets} are variables that represent a set of values. For example, $S_c$ corresponds to all the surface forms of the candidate entity $c$. The rest includes a set of candidate entities that are determined based on a specific position. \textbf{Functions} are applied on these variables as well as sets in order to define more detailed features.

We group the ranking features into four main categories as shown in Table~\ref{RankingFeaturesTable}. In the literature, the features used for disambiguation are basically divided into two main categories, namely local context-independent and global context-based features. In order to make it more explicit, we further break down the global features into three sub-categories. Those are mention-level, document-level and second stage features.

\begin{table}[h!]
\begin{center}
\scriptsize
\renewcommand\arraystretch{1.8}
\begin{tabular}{ll} \hline \hline
\bf Local Features : & \\
\textit{surface-form-edit-distance} &  dist($SF_c^{best}$, $SF_m$) \\ \cline{1-2}
\textit{entity-log-freq} & log($\sum_{s \in S_c}$ $freq$($SF_c^s$)) \\ \cline{1-2}
\textit{surface-from-type-in-binary} & binarize($T_c^{SF_c^{best}}$) \\ \cline{1-2}
\textit{entity-type-in-binary} & binarize($T_c^{e}$) \\ \cline{1-2}
\textit{$id^t$-typing-prob} & $P_c^n$, where $n \in$ \{\textit{Word},\textit{Surface},\textit{Synset},\textit{Brown},\textit{Entity}\} \\ \cline{1-2} 

\bf Mention-level Features : & \\
\textit{avg-surface-form-edit-distance} & avg( $\sum_{c' \in C^{m}}$ dist($SF_{c'}^{best}$, $SF_m$) ) \\ \cline{1-2}
\textit{max-diff-surface-form-log-freq} & $argmax_{c' \in C^m}$ $freq$($SF_{c'}$)-$freq$($SF_c$), if $freq$($SF_c$) is not max \\
& $freq$($SF_c$) - $argsecmax_{c' \in C^m}$ $freq$($SF_{c'}$), otherwise \\ \cline{1-2}
\textit{max-diff-doc-similarity} & $argmax_{c' \in C^m}$ cos($D_{c'}$,$D_t$)-cos($D_c$,$D_t$),if cos($D_c$,$D_t$) is not max  \\
& cos($D_c$,$D_t$) - $argsecmax_{c' \in C^m}$ cos($D_{c'}$,$D_t$), otherwise \\ \cline{1-2}
\textit{max-diff-$id^t$-typing-prob} & $argmax_{c' \in C^m}$ $P_{c'}^n$ - $P_c^n$, if $P_c^n$ is not max \\
& $P_c^n$ - $argsecmax_{c' \in C^m}$ $P_{c'}^n$, otherwise \\ \cline{1-2}

\bf Document-level Features : & \\
\textit{max-$id^t$-typing-prob-in-doc} & $argmax_{c' \in C_c^{t}}$ $P_{c'}^n$ \\ \cline{1-2}

\bf Second-stage Features : & \\
\textit{max-ranking-score} & $argmax_{c' \in C_c^{prev}}$ $R_{c'}$ if $C_c^{prev}$ is not empty \\ 
& $argmax_{c' \in C_c^{next}}$ $R_{c'}$ if $C_c^{next}$ is not empty \& $T_{c'}^{SF}=WikiID$ \\ 
& 0 , otherwise \\ \cline{1-2}
\textit{max-cos-sim-in-context} & $argmax_{c' \in C_{topNxM}}$ cos($E_c$, $E_{c'})$ * $P_{c'}^n$  \\
\hline \hline
\end{tabular}
\end{center}
\caption{\label{RankingFeaturesTable} List of Features used by the Ranking Model and their Descriptions.}
\end{table}

\textbf{Local Features} are those that consider the individual properties of the candidate without taking into account any other candidate. For example, the feature {\it surface-form-edit-distance} considers the edit distance between the most similarly matched surface form ($SF_c^{best}$) of the candidate in our Surface Form Data Set and the actual surface form ($SF_m$) of the mention. The more distant they are, the less likely the candidate is the actual one. In order to take into account the popularity of the candidate, the feature {\it entity-log-freq} uses the total number of times that named entity is seen in the Surface Form Data Set. It uses the logarithm to scale down and smooth the values. The more popular the candidate is, the more likely it might be mentioned. Another local feature {\it doc-similarity} is the cosine similarity between the doc2vec\footnote{We train Gensim's Doc2vec tool on the Wikipedia articles with embedding size=300 and set the other parameters to their default values in order to obtain document embeddings.} embeddings of the test document and the Wikipedia article of the candidate. The similarity between the two corresponds to the similarity between the context of the test document and the context in which the candidate is expected to be mentioned.

\begin{table}[h]
\begin{center}
\scriptsize
\renewcommand\arraystretch{1.05}
\begin{tabular}{cll} \hline \hline
& \bf Type & \bf Description \\ \hline
\multirow{6}{*}{\rotatebox{90}{Entity Types}} & & {\bf Entities labeled w/following YAGO synsets and BaseKB types} \\
& Person & wordnet\_person\_100007846, people.person, ...\\
& Organization & wordnet\_organization\_108008335, organization.organization, ... \\
& Location & wordnet\_site\_108651247, location.location, ...\\
& SportsTeam & wordnet\_team\_108208560, wordnet\_club\_108227214, ...\\
& Misc & {\it Any other dissimilar synset or type}\\ \hline
\multirow{13}{*}{\rotatebox{90}{Surface Form Types}} & & {\bf When ...} \\
& WikiID & $SF$ is same as the $Main Title$ of the entity\\
& Redirect & $SF$ is labeled as "redirect" in Wikipedia dump files \\
& Disambiguation & $SF$ is labeled as "disambiguation" in Wikipedia dump files \\
& FirstName & the entity is Person-type and $SF$ is a known first name (eg. John) \\
& Surname & the entity is Person-type and $SF$ is a known surname (eg. Smith)\\
& FirstWord & $SF$ is the first word in the main title of the entity \\
& LastWord & $SF$ is the last word in the $Main Title$ of the entity \\
& PrefixPhrase & $SF$ is prefix phrase in the $Main Title$ of the entity \\
& SuffixPhrase & $SF$ is suffix phrase in the $Main Title$ of the entity \\
& BeforeComma & $SF$ is the phrase before comma in the $Main Title$ of the entity\\ 
& OrgAcronym & $SF$ comprised of first letters of $Main Title$ of the entity \\
 \hline \hline
\end{tabular}
\end{center}
\caption{\label{EntityAndSurfaceFormTypes} Entity Types and Surface Form ($SF$) Types in SFDB.}
\end{table}
 
The two features {\it entity-type-in-binary} and {\it surface-form-type-in-binary} are binary vectors that encode the type of the candidate entity and the surface form, respectively. These types are based on the categorization of the entities and surface forms defined in Table~\ref{EntityAndSurfaceFormTypes}. We assign a type to each named entity in our KB based on whether it is labeled with one of the pre-defined YAGO synsets or BaseKB types. We define five main types. While \textit{Person}, \textit{Organization}, \textit{Location} and \textit{Misc} types are common in the named entity recognition literature, we added the \textit{SportsTeam} type since many mentioned entities in AIDA data sets are of this type. In case of the surface form types, we assign one or more applicable types to the $SF_c^{best}$. Other than \textit{Redirect} and \textit{Disambiguation} types in Table~\ref{EntityAndSurfaceFormTypes}, the rest is based on the similarity between the surface form and the $Main Title$ of the entity, which is obtained from its Wikipedia ID by replacing any underscore character with the blank space and discarding any phrase given in parenthesis. Certain combinations of these two types can give clues about the likelihood of the candidate. For example, it is more likely that mention of \textit{Person} type entity is made in surface form of type \textit{FirstWord} or \textit{Organization} type entity in \textit{OrgAcronym} type surface form.

 The final local feature \textit{$id^t$-typing-prob} is the probability ($P_c^n$) of labeling the mention with the pre-assigned cluster-based type of the candidate by the typing model $n$. More precisely, when the typing model is applied to the input text containing the mention, it outputs the probability of the mention belonging to each of the cluster-based types. As we know the pre-assigned cluster-based type of the candidate, the probability of that type is set as $P_c^n$. Since we have five different typing models, this feature contributes with five separate values. Note that each feature that includes $P_c^n$ contributes similarly.
 
\textbf{Mention-level Features} consider the relative value of an individual candidate's feature with respect to the other candidates' for the same mention ($C^m$). The first feature \textit{avg-surface-form-edit-distance} takes the average of all \textit{surface-form-edit-distance} feature values of $C^m$. Averaging helps us represent the edit distance of an average candidate. We can argue that the higher it is, the more likely that none of the candidates are the actual entity being mentioned. The features prefixed with \textit{max-diff-} compare the candidate's corresponding feature value with respect to the best value seen for the mention. The difference between the two values is used as the feature value. If the value is already the best, then it uses the second best value. The higher positive value it is, the more likely that the candidate is the actual one. Or, the lower negative value it is, the less likely that the candidate is the one. \textit{max-diff-} converts a context independent value to a context dependent by representing it with respect to the highest (or the next highest) seen value.

\textbf{Document-level Features} consider all occurrences ($C_c^t$) of the same candidate in the same test document. We have one such feature and it uses the highest seen $P_c^n$ value for the considered candidate. When the same entity is seen as a candidate in multiple mentions, each has its own $P_c^n$ value depending on the position of the mention in the document. The highest of them increases the chance of the other occurrences of the same candidate with the lower $P_c^n$ in the same document.

\textbf{Second-stage Features} are only available after the ranking model with the previously described features is trained and applied on the candidates once. Second-stage features use the ranking probability of each candidate ($R_c$). The first feature \textit{max-ranking-score} keeps the highest $R_c$ among the previous occurrences of the same candidate ($C_c^{prev}$) in the same document. If there is no previous occurrence of the candidate before the considered mention, then it looks at the future occurrences ($C_c^{next}$) only if $T_{c'}^{SF}$ of that candidate is \textit{WikiID}. This means that having an occurrence of the candidate with high $R_c$ increases the chance of the next occurrences of the same candidate in the same document. The same effect for the early occurrences can only happen if the future occurrence is mentioned in its full title. The second feature \textit{max-cos-sim-context} uses the cosine similarity between the entity embeddings ($E_c$) of the candidates\footnote{We use cooccurrence based entity embeddings obtained with the word2vecf tool. We set the window size to 5 and number of iterations to 20 on top of default settings.}. This feature is set to the highest cosine similarity value between the candidate and the highest ranked N candidates in the surrounding window of M mentions. 

\subsection{Neural Network Model for Disambiguation}
\label{Sec:RankerModel}

Our neural network for ranking candidate named entities is a two-layer feedforward neural network and one softmax layer\footnote{In our experiments, we observed that using the softmax layer instead of logistic regression in the PyTorch provides higher results for this binary classification task.} at the top. We use a dropout layer after each feedforward layer. We turn the ranking task into a binary classification task, where we classify each candidate as true or false candidate at the training time and then use the output probability of the true class to rank the candidates at the test time. For each candidate, we create its own input vector, which includes all numeric values for the features described in Section~\ref{Sec:RankingFeatures}. Since we do re-ranking, we have two disambiguation models in our experiments, the first one does the initial ranking and the second one does the final ranking based on additional ranking insight obtained from the first model. In both cases, we use the same network architecture, except the fact that the first model does not include the second-stage features. The values for those features are calculated from the output of the first model. Hence, in case of the test sets, we first need to run the first model on them and get the ranking scores for each candidate. After that, we can run the second model to get the final ranking results.

\section{Experimental Setup}

\subsection{Training and Test Data Sets}

\subsubsection{Data Sets for the Mention Typing}
\label{Sec:TypingModelDataSets}
We derived our data sets from Wikipedia which includes over five million well-written documents, each describing one named entity. Most of the mentions of named entities and concepts in the documents are manually annotated with an HTML anchor tag. However, Wikipedia Editing Guidelines suggest authors to annotate only the first occurrence of a named entity in the Wikipedia articles, which means that most of the mentions are not annotated in the articles. Hence, we try to auto-annotate the remaining occurrences in the documents. To do this, we look for the word sequences (with greedy match) that are previously annotated in the same document and then auto-annotate them at the rest of the document. This process increases the number of mentions considerably with an acceptable error rate. However, we did not try to annotate mentions with the partial names. 

\begin{table}[h]
\begin{center}
\scriptsize
\renewcommand\arraystretch{1.05}
\begin{tabular}{llc} \hline \hline
             & \bf Num. of   & \bf Average Num. of \\
\bf Data Set & \bf Instances & \bf Tokens per Context \\ \hline
WC-SmallTrain    & 981,000   & 22.2 \\ 
WC-LargeTrain    & 9,828,000 & 22.2 \\ 
WC-Test          & 9,700     & 22.2 \\
SFC-SmallTrain & 500,000   & 28.9 \\
SFC-LargeTrain & 5,000,000 & 28.9 \\
SFC-Test       & 50,000    & 29.0 \\
EC-SmallTrain    & 500,000   & 14.8 \\
EC-LargeTrain    & 5,000,000 & 14.8 \\
EC-Test          & 50,000    & 14.8 \\ \hline \hline
\end{tabular}
\end{center}
\caption{\label{PredictionDatasetsStats} Statistics on the Data Sets for the Mention Typing Models.}
\end{table}

After this pre-processing step, we create a separate data set for each context representation. Note that we represent the context in three different levels as described in Section~\ref{Sec:EntityEmbeddings}. In case of word-based context (WC), we break the Wikipedia articles into sentences and collect only those sentences that have at least one named entity mention and the length is between 10-50 words. We use the sentence boundary detector tool and tokenizer in the Stanford CoreNLP \cite{Manning14}. We ended up having 46.8M such sentences. Out of those, we randomly selected sentences and created WC-* data sets. In case of surface form-based context (SFC), we start with the same sentences and then for each mention, we get the mention surface forms of the previous and next 10 mentions in the same document. Again we randomly selected instances and created SFC-* data sets. For the entity-based context (EC), around each mention we collected the previous and next 10 named entities in the same document. This is called EC-* data sets. Sample training instances are already exemplified in Figure~\ref{fig:ExmapleTypingModelTrainingData} in Section~\ref{Sec:PreparingTrainingData}. For each type, we created small and large training sets and a test set as given in Table~\ref{PredictionDatasetsStats}. The number of instances and the average number of tokens in the context (left and right combined) are also provided. 

\subsubsection{Named Entity Disambiguation Data Sets}

There are a number of publicly available data sets with different characteristics for the NED task. For training, development and test purposes, we use \textbf{AIDA} \cite{Hoffart11}, which is derived from Reuters news articles of the CoNLL 2003 NER task. Being the widely used and largest NED data set, it comes in three pieces: AIDA.train (train), AIDA.testa (dev), and AIDA.testb (test). We report the results on AIDA-testb as in-domain evaluation results, since all our training is done on AIDA-train. In order to see how the model that is trained on AIDA.train achieves on the data sets that exhibit different characteristics, we test that model on a number of other test sets. This is called cross-domain evaluation. For this purpose, we use \textbf{MSNBC} \cite{Cucerzan07}, \textbf{AQUAINT} \cite{Milne08} and \textbf{ACE2004} \cite{Ratinov11}, which are the next three most frequently used test sets. ACE2004 is a subset of ACE2004 Coreference documents, while AQUAINT contains news articles from the Xinhua, New York Times and Associated Press. Since they are also used for wikification, they include Wikipedia concepts apart from named entities. As the recent studies used the revised versions of these data sets prepared by Guo and Barbosa \shortcite{Guo16}, we report our results on these revised versions as well. Apart from these, we also consider three more test sets in order to observe how our system performs on shorter documents. \textbf{KORE50} \cite{Hoffart12} includes 50 short sentences on various topics such as celebrities, music etc. Most of the mentions are single-word such as first names, which makes the deduction of the actual mentioned named entity very difficult. \textbf{RSS-500} \cite{Roder14} includes short formal text collected from a data set containing RSS feeds of the newspapers compiled in \cite{Goldhahn12}. The texts are on a wide range of topics such as world, business, science etc. \textbf{Reuters-128} \cite{Roder14} is a small subset of the well-known Reuters-21587 corpus containing short news articles about the economy.

\begin{table}[h]
\begin{center}
\scriptsize
\renewcommand\arraystretch{1.05}
\begin{tabular}{lccccc} \hline \hline
\bf Data Set & \bf Domain & \bf RefKB & \bf \#Mentions & \bf Avg\#M/D & \bf Avg\#W/S \\ \hline
AIDA.train                  & news & Wikipedia & 18,448 & 19.5 & 15.7 \\ 
AIDA.testa (dev)            & news & Wikipedia & 4,791 & 22.2 & 17.2 \\
AIDA.testb (test)           & news & Wikipedia & 4,485 & 19.4 & 14.5 \\
MSNBC                       & news & Wikipedia & 656 & 32.3 & 28.8 \\
MSNBC\textsuperscript{rev}  & news & Wikipedia & 655 & 32.8 & 27.5 \\
AQUAINT                     & news & Wikipedia & 730  & 14.3 & 28.5 \\
AQUAINT\textsuperscript{rev}& news & Wikipedia & 722 & 14.4 & 28.5 \\
ACE2004                     & news & Wikipedia & 255 & 4.5 & 37.0 \\ 
ACE2004\textsuperscript{rev}& news & Wikipedia & 256 & 4.5 & 38.6 \\ 
KORE50                      & mixed & DBpedia  & 143 & 2.9 & 14.6 \\
RSS-500                     & RSS-feeds & DBpedia & 517 & 1.0 & 32.2 \\ 
Reuters-128                 & news & DBpedia & 621 & 4.9 & 28.0 \\ \hline \hline
\end{tabular}
\end{center}
\caption{\label{TestSetStats} Statistics on the Disambiguation Data Sets.}
\end{table}

Details about these data sets are given in Table~\ref{TestSetStats}. Note that since we use Wikipedia as our reference KB, we map the DBpedia-based annotations in KORE50, RSS-500 and Reuters-128 to the corresponding Wikipedia-based IDs. The AIDA sets have the most annotations among all, which makes them suitable for training and  development. KORE50, RSS-500, ACE2004, and Reuters-128 are the sets that have the smallest average number of mentions per document (Avg\#M/D). In terms of sentence-based context size, the KORE50 and AIDA sets have the smallest average number of words per sentence (Avg\#W/S). Some of these data sets include NIL annotations, which have no corresponding named entity at the reference KB. As a design decision, we exclude such annotations and report the results accordingly in our experiments.

\subsection{Evaluation Metrics}

In this study, we report the standard micro-average F1-score for the evaluation of the cluster prediction model and the disambiguation model. F1-score is the harmonic mean of the precision and recall measures. Precision measures the percentage of machine's predictions that are correct compared to the gold (human) annotations. On the other hand, recall measures the percentage of the gold annotations that are predicted correctly by the machine. In case of the disambiguation results, a number of studies reported their results in bag-of-title (BoT) F1-score \cite{Milne08,Ratinov11}, which is designed for Wikification systems. It is used to evaluate a NED system for indexing purposes. Hence, it discards the duplicate gold annotations and predictions in a document and uses the same F1-score metric on the filtered out numbers. Moreover, we report our results in InKB accuracy in case of the AIDA test set due to the convention in the literature. On the other test sets, we use a threshold on the ranking probability and do not assign an entity to a mention if our system is not confident with the assignment (i.e., the ranking probability is below the threshold). In our experiments, we use 0.03 as the threshold.

\subsection{Optimizing Clustering for Better Disambiguation}
\label{Sec:ClusterOptimization}

Important thing before getting into the experimental results is tuning of the clustering for better disambiguation. As described in Section~\ref{Sec:EntityClustering}, how we cluster named entities determines the effect of the discriminative power of clustering on the disambiguation step. The more clusters we use, the fewer entities each cluster has, hence the more discriminative the clustering becomes. For example, assume that we have a list of candidates for a mention and this set of candidates are grouped together based on their assigned cluster-based types. Think of one particular group that contains the gold standard entity. All candidates in that group are given the same chance of being the actual referred entity as they all have the same cluster-based type. Therefore, when we have fewer entities in that group, we expect it to be relatively easier to disambiguate the correct entity among all candidates compared to when we have more entities in that group On the other hand, when we have more clusters, this makes the typing model difficult to solve due to the increase in the number of classes (i.e. types). Having said all these, we look for an optimization to deal with this trade-off.

There are two main factors we considered for optimization. The first is the window size chosen while calculating the entity embeddings on which we run the K-means algorithm. The second and more important factor is the number of clusters to be generated by the clustering algorithms. However, it is not feasible to try all possible combinations and measure which one achieves the best performance at the disambiguation step. That is because for each combination, we need to get the clusterings, train their typing models, train the ranking model on their predictions and then finally measure the disambiguation success. Instead, we introduce the following three-step approach.

The first step is to produce a wide range of clusterings for the named entity space with different window sizes for embeddings and different numbers of clusters. The second step includes selecting a good sample of clusterings and training a typing model with each selected clustering. Since we use multiple mention typing models for disambiguation, the third step is to select one clustering for each typing model so as to achieve high at the disambiguation step. In order to choose this combination, we also propose a heuristic that minimizes the number of times we train the ranking model. Below, we explain each step in more detail.

The first step involves creating various clusterings of named entities for each of the five clustering approaches. The first and primary parameter is the number of clusters. In case of the \textit{Brown}-based approach, we run the tool with a number of clusters from 1000 to 1500 in 100 incremental steps. For the other four clustering approaches, we run the K-means algorithm with number of clusters from 600 to 3000 in 200 incremental steps. The second parameter is related to the calculation of the entity embeddings which are only used by K-means. In case of \textit{Synset}-based approach, since there is no sequential context, we use word2vecf and the only parameter we change is the \textit{iteration}. For the other three approaches, we use \textit{word2vec} and experiment with different values of the \textit{window} parameter. We use window sizes of 2 and 3. Using other values does not provide any better results. We set the \textit{iteration} parameter for the \textit{Synset}-based approach to 20. At the end of the first step, we get a total of 70 clusterings.

At the second step, we pick one or more parameter combinations for each clustering approach. However, instead of selecting randomly, we proposed a metric that helps us evaluate each clustering based on how well it might help discriminate the candidates for the benefit of the disambiguation model. This metric is called Average Gold Candidate Cluster Size (AGCCS). When we group the candidates of the mention based on the clustering at hand, we end up having clusters of candidates, or candidate clusters. The one that holds the gold (true) candidate is called the gold candidate cluster. The smaller the size of the gold candidate cluster is, the less number of candidates are being favored by the cluster prediction model at the disambiguation step. In other words, it becomes more discriminative. For each parameter combination, we calculated AGCCS value on the AIDA.train data set, which includes over 18000 mentions with an average of 84 candidates for each mention.

\begin{figure}[h]
\centering
\includegraphics[width=120mm]{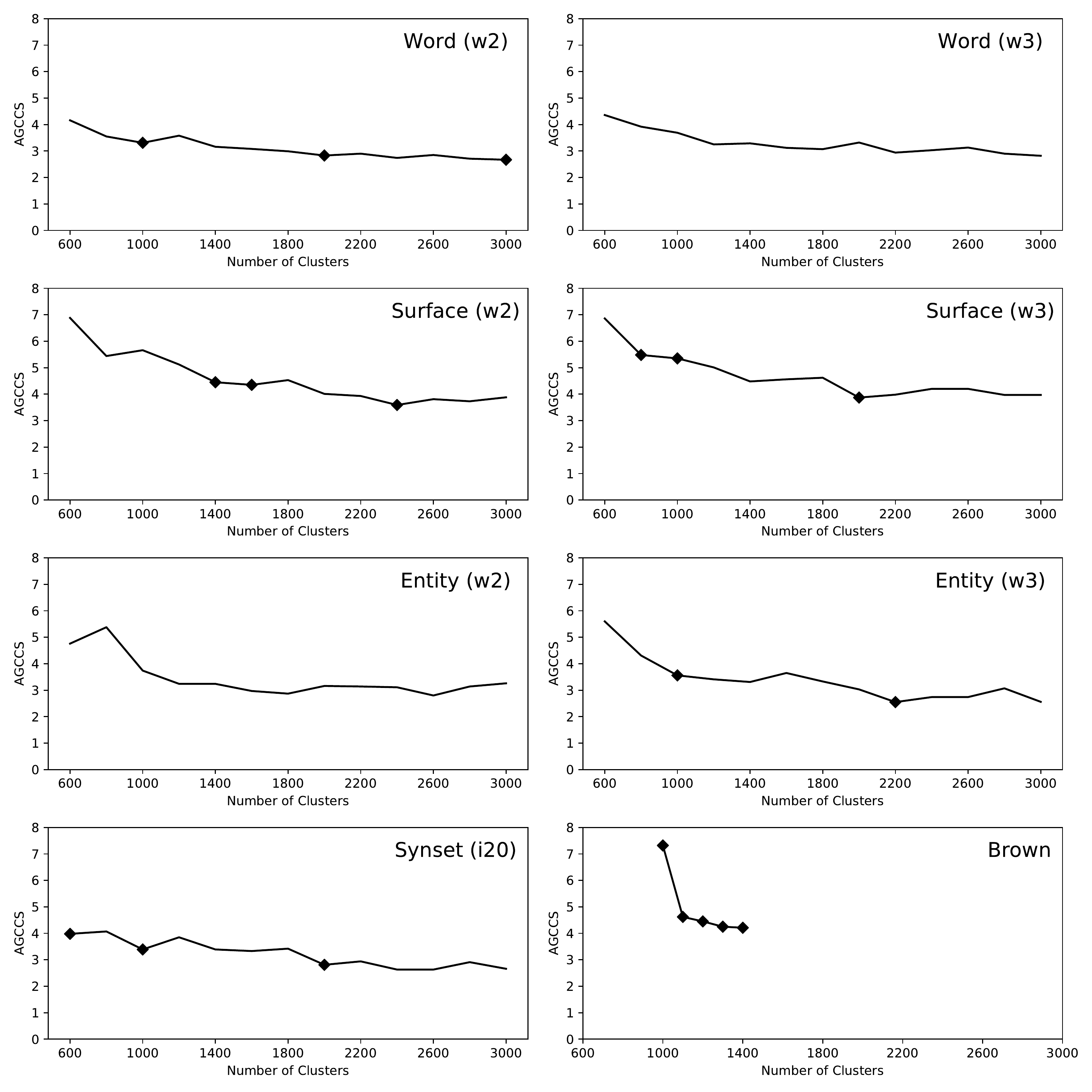}
\caption{Change of Average Gold Candidate Cluster Size (AGCCS) as the Number of Clusters is Increased for each Clustering Approach.}
\label{fig:AGCSSTable}
\end{figure}

Figure~\ref{fig:AGCSSTable} shows how AGCCS changes, as we increase the number of clusters. Generally, it drops, because using more clusters leads to less number of entities inside the clusters, hence the smaller size for the gold candidate cluster. Note that AGCCS generally decreases fast at first but does not drop below 2.5. The dots in Figure~\ref{fig:AGCSSTable} show the chosen cases that we train a typing model for. We particularly pick the ones with the lowest AGCCS value. Moreover, we also consider some other local minimums, since they are obtained with less number of clusters. At the end, we have 19 different typing models calculated\footnote{We have three \textit{Word} models for 1000, 2000, 3000 clusters with window=2; six \textit{Surface} models for 1400, 1600 and 2400 clusters with window=2 and 800, 1000 and 2000 clusters with window=3; two \textit{Entity} models for 1000 and 2200 clusters with window=3; three \textit{Synset} models for 600, 1000, and 2000 clusters with iter=20; and five \textit{Brown} models for 1000, 1100, 1200, 1300, and 1400 clusters}, hence 19 dots in Figure~\ref{fig:AGCSSTable}.

The third step is to pick the best clustering combination. Note that based on 19 calculated typing models, we ended up having 540 different combinations\footnote{540 = 3 \textit{Word} x 6 \textit{Surface} x 3 \textit{Synset} x 6 \textit{Brown} x 2 \textit{Entity}.}. However, it is not feasible to train a disambiguation model for each combination and pick the best performing one. Instead, we proposed a heuristic that scores the typing model combinations according to  Equation~\ref{eq:TypingModelSelectionHeuristic} . We omit the \textit{Entity} typing model in order not to involve the second-stage ranking into this procedure. After getting the first-stage fixed, we pick the best performing \textit{Entity} model which is obtained with 2200 clusters.

\begin{equation}
\centering
\begin{aligned}
\argmin_{w \in Word, s \in Surface, y \in Synset, b \in Brown}  P^w + P^s + P^y + P^b \\
\textrm{where} \quad P^t = \sum_{m \in M} \sum_{c_{ng} \in \{C^m - c_g\} } P_{c_{ng}}^t - P_{c_g}^t, \quad \textrm{if} \quad P_{c_{ng}}^t > P_{c_g}^t
\end{aligned}
\label{eq:TypingModelSelectionHeuristic}
\end{equation}

Note that a typing model $t$ outputs the probabilities of the cluster-based types for a given $mention$. Since we know the pre-assigned cluster-based type of each candidate $c_{type}$, we also know $P(c_{type}|mention)$, or $P_c^t$ in short. The ideal case from the disambiguation point of view is to have all typing models assign the highest probability to the gold candidate $c_g$ so that it can be easily distinguishable from the rest of the candidates (i.e. $C^m - c_g$), which are called non-gold candidates, or $c_{ng}$. However, in a real world scenario, typing models can make a classification mistake which leads to $c_{ng}$ having higher probability than $c_g$. We call them competing non-gold candidates. Having higher typing model probability falsely favors them in the disambiguation step. Having said that, Equation~\ref{eq:TypingModelSelectionHeuristic} chooses the model combination such that it minimizes the aggregate probability difference between gold candidate and competing non-gold candidates calculated over all mentions $M$. For these calculations, we use the AIDA.train data set.

At the third step, instead of picking the typing model combination that has the lowest value according to Equation~\ref{eq:TypingModelSelectionHeuristic}, we select the lowest scored $10$ combinations and train a ranking model for each. Then, we score them on the AIDA.testa development set and pick the one that achieves the highest disambiguation score. The selected combination is \textit{Word} model with 1000 clusters, \textit{Synset} model with 1000 clusters, \textit{Brown} model with 1300 clusters, and \textit{Surface} model with 2400 clusters. For the \textit{Entity} model in the second-stage of the ranking, we use the model with 2200 clusters.

\section{Experimental Results}

\subsection{Evaluation of the Candidate Generator}

In order to evaluate the candidate generator, we use the gold recall measure, which is defined as the percentage of the annotated (gold) named entities in the data set that have been suggested by the candidate generator. Table~\ref{CandidateGeneratorRecalls} gives the gold recall values for our candidate generator and two other candidate generators in the literature. Note that most of the NED studies use a smaller number of candidates in order to discard the least possible cases before doing the ranking. Hence, \textit{Ratinov} \cite{Ratinov11} and \textit{Ganea} \cite{Ganea17} use the highest scored 20 and 30 candidates per mention, respectively. In our experiments, we use top 100 ($N=100$) candidates in order to increase the recall of the NED system and to have more negative examples during the training of the model. To evaluate our candidate generator, we also calculate gold recall for top 20 and 30 candidates. Moreover, in order to measure the contribution of the second stage of our candidate generation algorithm described in Section~\ref{Sec:CandidateGenerator}, we also calculate these recalls without applying the second stage. The results of the proposed candidate generator as well as the ones of \cite{Ratinov11} and  \cite{Ganea17} are given in Table~\ref{CandidateGeneratorRecalls}. In addition to these, Hakimov \textit{et al.} \shortcite{Hakimov16} reported a recall of 97.7 on AIDA.train when $N=100$.

When we compare our recall values with other studies, in almost all cases our proposed candidate generator achieves better performance except with respect to the performance of Ratinov \textit{et al.} \shortcite{Ratinov11} on the AQUAINT. The results also show that the second stage in the candidate generator produces a substantial improvement. It means that using candidates from similarly titled surrounding mentions and extending the candidates even further with their cooccurring entities from Wikipedia results in higher recall values. The only exception is with the AQUAINT and AQUAINT\textsuperscript{rev}.

\begin{table}[h]
\begin{center}
\scriptsize
\renewcommand\arraystretch{1.1}
\begin{tabular}{lccccccccc} \hline \hline
            & \multicolumn{3}{c}{\bf Ours w/o 2nd Stage} & \multicolumn{3}{c}{\bf Ours w/ 2nd Stage}  &\bf Ganea & \bf Ratinov \\
\bf Data Set&  N=100 & N=30 & N=20 & N=100 & N=30 & N=20 & N=30   &  N=20 \\ \hline 
AIDA.train                  & 97.97 & 97.56 & 97.16 & 99.74 & 99.22 & 98.61 & -    & -     \\ 
AIDA.testa                  & 97.63 & 97.26 & 97.07 & 99.85 & 99.22 & 98.24 & 96.6 & -     \\
AIDA.testb                  & 97.87 & 97.07 & 96.58 & 99.62 & 98.59 & 97.36 & 98.2 & -     \\
MSNBC                       & 91.63 & 91.63 & 91.01 & 99.70 & 99.22 & 98.29 & 98.5 & 88.67 \\
MSNBC\textsuperscript{rev}  & 91.91 & 91.91 & 91.30 & 99.85 & 99.08 & 98.47 & -    & -     \\
AQUAINT                     & 96.93 & 96.23 & 95.25 & 97.40 & 96.09 & 94.69 & 94.2 & 97.83 \\
AQUAINT\textsuperscript{rev}& 97.92 & 97.37 & 96.68 & 98.06 & 96.95 & 96.12 & -    & -     \\
ACE2004                     & 96.86 & 95.29 & 93.73 & 96.86 & 95.65 & 94.51 & 90.6 & 86.85 \\
ACE2004\textsuperscript{rev}& 96.88 & 95.70 & 94.14 & 96.88 & 96.09 & 94.92 & -    & -     \\
KORE50                      & 86.01 & 82.52 & 81.82 & 92.31 & 88.81 & 87.41 & -    & -     \\
RSS-500                     & 88.39 & 86.27 & 85.69 & 89.56 & 87.43 & 86.85 & -    & -     \\
Reuters-128                 & 88.41 & 87.60 & 87.28 & 95.65 & 90.82 & 89.21 & -    & -     \\ \hline \hline
\end{tabular}
\end{center}
\caption{\label{CandidateGeneratorRecalls} Gold Recall Values for Candidate Generation on the NED Data Sets.}
\end{table}

When we perform error analysis, we observe that in the case of KORE50, small context and mostly one word mentions result in low recall. In case of RSS-500, the annotated mentions only include part of the existing surface form (e.g., only the word ``France'' is annotated for the available surface form ``Tour de France''). Since we do not take into account the immediate surrounding words of the annotated mentions during the search or use an off-the-shelf named entity recognizer \cite{Phan17} to expand the boundaries, we achieve relatively low recall on RSS-500. Table~\ref{CandidateGeneratorStats} reports the average number of candidates per mention generated for $N=100$. It also shows the average length of the mentions in characters  and the average edit distance between the surface form ($SF_m$) of the mention and the best matched surface form of the candidate ($SF_c^{best}$)  per candidate. The highest values for both metrics are seen in Reuters-128, while the average mention length values on the AIDA sets also show their characteristic difference from the other sets, which is important considering that we train our disambiguation model on the AIDA.train and test it on the other sets.

\noindent
\begin{table}[h]
\begin{center}
\scriptsize
\renewcommand\arraystretch{1.1}
\begin{tabular}{lccc} \hline \hline
             & \bf Average Num.       & \bf Average      & \bf Average \\
\bf Data Set & \bf of Candids (N=100) & \bf Mention Length & \bf Edit Distance   \\ \hline 
AIDA.train                  & 84.65 & 8.9 & 1.7\\ 
AIDA.testa                  & 84.35 & 9.0 & 1.8\\
AIDA.testb                  & 84.45 & 8.9 & 1.9 \\
MSNBC                       & 92.97 & 10.2 & 2.6 \\
MSNBC\textsuperscript{rev}  & 94.22 & 9.9 & 2.5 \\
AQUAINT                     & 78.07 & 11.6 & 2.1 \\
AQUAINT\textsuperscript{rev}& 78.22 & 11.6 & 2.1 \\
ACE2004                     & 79.15 & 11.0 & 2.2 \\
ACE2004\textsuperscript{rev}& 79.17 & 11.0 & 2.2 \\
KORE50                      & 85.44 & 6.3 & 1.2 \\
RSS-500                     & 70.90 & 11.3 & 2.7 \\
Reuters-128                 & 72.79 & 12.6 & 3.8 \\ \hline \hline
\end{tabular}
\end{center}
\caption{\label{CandidateGeneratorStats} Statistics on the Candidates Generated for  each NED Data Set.}
\end{table}

\subsection{Contribution of Specialized Word Embeddings in Mention Typing}

Instead of using regular word embeddings (i.e. $W_R$) as input for our mention typing models, we introduced two different word embeddings in Section~\ref{Sec:SpecializedWordEmbeddings}. Those are the cluster-centric word embeddings ($W_{CC}$) and surface form based word embeddings ($W_{SF}$). $W_{CC}$ is proposed as an optimized version of $W_R$, as it is influenced by the applied clustering during the calculation. Both embeddings are obtained with word2vec\footnote{We set  window=2, size=300, and use the default values for the other parameters.} on all Wikipedia articles. They are used as an input to $Left$, $Right$ and $Surface_1$ components (either LSTM or CNN layers shown in Figures~\ref{fig:LSTMPredictionModel} and \ref{fig:CNNPredictionModel}) of the typing model. In case of $W_{SF}$, it is obtained with word2vecf\footnote{We set size=300, and use the default parameter values for the other parameters.} from the large surface form data set as described in Section~\ref{Sec:SpecializedWordEmbeddings}. Different from the previous two embeddings, $W_{SF}$ is used as an input to the $Surface_2$ component of the typing model. Note that, like $Surface_1$ component, $Surface_2$ also takes the words of mention's surface form as input but in $W_{SF}$ embeddings instead.

In order to measure the contribution of using $W_{CC}$ over $W_R$ and using additional $W_{SF}$, we trained both LSTM\footnote{We set the learning rate to 0.1 and use the standard gradient descent optimizer with Nesterov momentum of 0.9. We set the weight decay rate to 1.2e-06 and clip the gradients at 2.0. We set the hidden state size to 600 for all LSTM layers. Wherever applied, dropout probability is set to 0.5. We use batch size of 200.}- and CNN\footnote{We use the same parameter values as in the LSTM-based model, except the clip value is set to 1.0. The CNN filter sizes are set to 64.}-based models with different embedding combinations for each typing model. Note that each typing model is trained and tested on its own designated data set described in Section~\ref{Sec:TypingModelDataSets}. To consider all combinations in a feasible time frame, the small version of those data sets (the ones named *-SmallTrain and *-SmallTest) are used. In Table~\ref{CCWE_SFWE_Contributions_Table}, we measured how accurately the typing model predicts the cluster-based type labels assigned to each mention in the test sets. The results are given in F1 score.

\begin{table}[h]
\begin{center}
\scriptsize
\renewcommand\arraystretch{1.1}
\begin{tabular}{lllll|ccccc} \hline \hline
\bf Arch. & \multicolumn{4}{c}{Inputs} & \multicolumn{5}{c}{Mention Typing Models} \\
\bf Type & \bf Left & \bf $Surface_1$ & \bf $Surface_2$ & \bf Right &\bf \textit{Word} &\bf \textit{Synset} &\bf \textit{Surface} &\bf \textit{Brown} &\bf \textit{Entity}  \\ \hline
LSTM & $W_{R}$ & $W_{R}$ & \textit{NotUsed} & $W_{R}$      & 67.5 & 66.6 & 66.1 & 68.6 & 67.1 \\ 
LSTM & $W_{CC}$& $W_{CC}$& \textit{NotUsed} & $W_{CC}$     & +2.4 & +1.9 & +5.3 & +4.4 & +0.9 \\ 
LSTM & $W_{R}$ & $W_{R}$ & $W_{SF}$ & $W_{R}$ & +4.6 & +1.0 & +6.4 & +4.8 & +2.2 \\ 
LSTM & $W_{CC}$& $W_{CC}$& $W_{SF}$ & $W_{CC}$& +5.3 & +3.0 & +7.0 & +6.3 & +2.5 \\ \hline
CNN  & $W_{R}$ & $W_{R}$ & \textit{NotUsed} & $W_{R}$      & 69.1 & 61.1 & 58.3 & 60.3 & 61.4 \\
CNN  & $W_{CC}$& $W_{CC}$& $W_{SF}$ & $W_{CC}$& +3.7 & +7.5 & +14.5& +10.6& +6.3 \\ \hline \hline
\end{tabular}
\end{center}
\caption{\label{CCWE_SFWE_Contributions_Table}Showing the Contribution of the $W_{CC}$ (over $W_R$) and $W_{SF}$ Embeddings When Typing Models are Trained and Tested on *-SmallTrain and *-SmallTest Sets.}
\end{table}

The first rows for the LSTM and CNN-based models in Table~\ref{CCWE_SFWE_Contributions_Table} consider the case where we do not use any special word embedding except the $W_{R}$. The following rows show how much the F1-scores change with respect to the first row (i.e., the baseline), first when we replace $W_{R}$ with $W_{CC}$, then when we use $W_{SF}$. Shown at the last rows, using both special embeddings together increases the scores considerably, up to 3 to 7 points. Their contribution is even more visible in case of the CNN-based models. When we do the same analysis on the other NED sets, we also observe very similar contribution levels. Based on these results, we use $W_{CC}$ and $W_{SF}$ together in the rest of our experiments.

\subsection{Cluster-based Mention Typing Results}

In order to evaluate the cluster-based mention typing models, we train and test our five different models on the *-LargeTrain and *-LargeTest sets defined in Section~\ref{Sec:TypingModelDataSets}. The results in F1-score are given in Table~\ref{ModelResults} for both the LSTM- and CNN-based models. However, since they are tested on different sets, the F1-scores are not comparable\footnote{In order to keep the results as much comparable as possible, we train all typing models based on the same number of clusters, which is 1000.} across the table. Hence, the average loss per instance is included in parenthesis. It is calculated by normalizing the total cross entropy loss value given by the model on the test set with the number of instances in that set. Note that average loss per instance is a better evaluation metric than F1-score for assessing the quality of the predictions for the disambiguation step, since the prediction probabilities are used as features in the ranking model. F1-score only measures how accurate the model predicts the true label, whereas average loss per instance indirectly takes the model's probability of that predicted label into account.

\begin{table}[h]
\begin{center}
\scriptsize
\renewcommand\arraystretch{1.1}
\begin{tabular}{lccccc} \hline \hline
\bf Arch. Type      &  \bf \textit{Word}  & \bf \textit{Synset}  & \bf \textit{Surface}  & \bf \textit{Brown} & \bf \textit{Entity}    \\ \hline
LSTM & 81.2 (1.10) & 78.6 (1.52) & 83.4 (0.80) &  83.8 (0.68) &  82.1 (0.74) \\ 
CNN  & 81.0 (1.09) & 79.0 (2.14) & 81.0 (0.83) &  81.6 (0.75) &  79.6 (1.43) \\ \hline \hline 
\end{tabular}
\end{center}
\caption{\label{ModelResults}F1-scores and Average Loss per Instance Values for the Mention Typing Models Trained and Tested on the *-LargeTrain and *-LargeTest Sets.}
\end{table}

When the context is local as in the \textit{Word} and \textit{Synset} models, CNN performs similarly to LSTM. However, in general the LSTM-based models outperform the CNN-based models. This is supported by both F1-score and average loss per instance values. Hence, in the rest of our experiments, we use LSTM for all mention typing models. 

When we compare the different models with each other through the average loss per instance value, the \textit{Synset} model turns out to be the worst performer and the \textit{Word} model comes after that. This means that these two sentence-based models are outperformed by the document-level typing models, namely \textit{Surface}, \textit{Brown}, and \textit{Entity}. This might be expected due to the larger context at the document-level. In case of the worst performer, the synset-based model is based on clustering of the entity embeddings that have not originated from the context, but are based only on the similarity of the assigned synsets. In other words, the cluster-based type labels used for training the \textit{Synset} model are not optimized for the contextual similarity. This might affect the success of the typing model which learns to predict based on the contextual similarities. The \textit{Synset} model not being optimal for the typing model is also supported by the results in Table~\ref{CCWE_SFWE_Contributions_Table}. Specialized word embeddings do not help the \textit{Synset} model much compared to the other models.

\subsection{Disambiguation Results}

For the disambiguation step, we trained our disambiguation model\footnote{We set the learning rate to 0.05 and use the standard gradient descent optimizer with Nesterov momentum of 0.9. The first and the second layers contain 500 and 300 units, respectively. The dropout values after the first and the second layers are set to 0.1 and 0.7, respectively. The training data are given in batches of 400 instances.} on the AIDA.train data set and did the feature selection based on the AIDA.testa development set. We train our ranking model in two-stages. At the first stage, the model is trained and run on all data sets. The ranking probabilities for the candidates are stored and used in the additional features, which are classified as the second-stage features in Section~\ref{Sec:RankingFeatures}. Our final results\footnote{We train our system 20 times with different seed values.} are obtained after training the ranking model with all the features. The results are reported in Table~\ref{NEDResultsOnLargeDocs} (for test sets with large context) and Table \ref{NEDResultsOnShortDocs} (for test sets with small context). 

\begin{table}[h]
\begin{center}
\scriptsize
\renewcommand\arraystretch{1.1}
\begin{tabular}{lccccccc} \hline \hline
\bf System               & MSN & MSN\textsuperscript{rev} & AQU & AQU\textsuperscript{rev} & ACE & ACE\textsuperscript{rev}  & AIDAb  \\ \hline
Phan {\it et al.} 2017                         & 91.8 &  -   &   -  &  -   & 92.9 &  -   & -    \\ 
Ganea and Hofmann 2017                         &  -   & 93.7 &  -   & 88.5 &  -   & 88.5 & 92.2 \\ 
Guo and Barbosa 2016                           & -    & 92.0 &  -   & 87.0 &  -   & 88.0 & -    \\
Yamada {\it et al.} 2016                       & -    & -    &  -   &  -   &  -   &  -   & 93.1 \\
Phan {\it et al.} 2018                         & 91.0 &  -   & 87.9 &  -   & 88.3 &  -   & -    \\
Sil {\it et al. } 2018                         & -    & -    &  -   &  -   &  -   &  -   & 94.0 \\
Le and Titov 2018                         & -    & 93.9 &  -   & 88.3 &  -   & 89.9 & 93.1 \\
Radhakrishnan {\it et al. } 2018          & -    & -    &  -   &  -   &  -   &  -   & 93.0 \\
Raiman and Raiman 2018                    & -    & -    &  -   &  -   &  -   &  -   & 94.8 \\
Fang {\it et al. } 2019                   & -    & 92.8 &  -   & 87.5 &  -   & 90.5 & 94.3 \\
Liu {\it et al. } 2019                    & -    & -    &  -   & 87.3 &  -   & 86.6 & 87.6 \\[-1ex] \hline
\textit{Cheng and Roth 2013 [BoT]}                      & \textit{90.0} &  -   & \textit{90.0} &  -  & \textit{86.0} &  -   & -    \\
\textit{Yang et al. 2018 [BoT]}                & -    & \textit{92.6} &  -   & \textit{90.5} &  -   & \textit{89.2} & \textit{95.9} \\[-1ex] \hline \hline
Ours                                           &      &      &      &      &      &      &      \\ 
w/o Typing Models (Baseline)                   & 87.3 & 88.4 & 84.5 & 87.0 & 80.8 & 82.0 & 81.4 \\ 
w/4 T.Models (1\textsuperscript{st} Stage)     & 91.9 & 92.6 & 88.4 & 90.7 & 89.5 & 90.3 & 89.9 \\ 
w/5 T.Models (2\textsuperscript{nd} Stage)     & \bf 92.6 & 93.0 & \bf 89.0 & \bf 90.7 & 90.0 & \bf 91.1 & 93.2 \\[-1ex] \hline
\textit{w/5 T.Models (2\textsuperscript{nd} Stage)} [BoT] & \textit{90.8} & \textit{92.1} & \textit{89.8} & \textit{91.9} & \textit{91.8} & \textit{93.2} & \textit{92.6}     \\[-1ex] \hline \hline
\end{tabular}
\end{center}
\caption{\label{NEDResultsOnLargeDocs}Results in F1 and BoT F1 on the NED Test Sets with Large Context.}
\end{table}

The upper part of the tables provides the results reported by the previous studies in the literature. Note that the results from the two studies at the bottom are given in BoT F1-score. The lower part of the tables presents our results given in traditional micro F1-score and BoT F1-score. The first line in "Ours" part shows the results taken without using any features related to the mention typing models or the second stage. We call it our baseline. As we add the features calculated with the four typing models (namely \textit{Word}, \textit{Synset}, \textit{Surface}, and \textit{Brown}), we can observe an increase of 2 to 8 points on all test sets. Next, we apply the second-stage, where we add the second stage features and the features calculated with the \textit{Entity} typing model. The results are improved further, especially on AIDA.testb with 3.3 points increase. 

In order to compare our results with the SOTA results in Table~\ref{NEDResultsOnLargeDocs}, we perform the randomization test with respect to the studies that achieved close to our results. On AQUAINT, we achieve better than Phan \textit{et al.} (2018) at statistically significant level\footnote{We compared our 20 runs with the one run of them that produced the same reported F-score and observed p-value=0.014 on average.}. On MSNBC, our higher F1-score turned out to be not statistically significant compared to Phan \textit{et al.} (2017). On the revised ACE2004, our higher result is not statistically significant\footnote{Table 10 includes the adjusted F1-score as they did not output prediction for 5 mentions of 256 that we have. Their reported F1-score of 91.2 is on 251 mentions, on which we achieve an F1-score of 91.7.} compared to Fang \textit{et al.} (2019). On the revised AQUAINT, our results are 2.2 points higher than Ganea and Hoffman (2017) and 1.4 points higher in BoT F1 with respect to Yang \textit{et al.} (2018). However, we are unable to perform valid randomization tests\footnote{Ganea and Hoffman (2017) provided the output of one run. The F1-score calculated on that run is 91.1, while the score reported in their study as the average of five runs is 88.5 $\pm$ 0.4. The provided output might be from their best run, while the score from our best run is 91.5.} with respect to their results.

\begin{table}[h]
\begin{center}
\scriptsize
\renewcommand\arraystretch{1.1}
\begin{tabular}{lccc} \hline \hline 
\bf System                                     & KORE50 & RSS-500 & Reuters-128 \\ \hline
Phan {\it et al. } 2017                        &  79.4  &  80.4   &  91.8       \\ 
Phan {\it et al. } 2018                        &  78.7  &  82.3   &  85.9       \\ \hline \hline
Ours                                           &        &         &             \\
w/o Typing Models (Baseline)                   &  40.4  &  68.9   &  72.0       \\ 
w/4 T.Models (1\textsuperscript{st} Stage)     &  56.1  &  74.4   &  76.5       \\ 
w/5 T.Models (2\textsuperscript{nd} Stage)     &  57.5  &  76.1   &  79.3       \\[-1ex] \hline
\textit{w/5 T.Models (2\textsuperscript{nd} Stage)} [BoT] &  \textit{58.5}  &  \textit{77.7}   &  \textit{88.6}       \\[-1ex] \hline \hline
\end{tabular}
\end{center}
\caption{\label{NEDResultsOnShortDocs}Results in F1 and BoT F1 on the NED Test Sets with Short Context.}
\end{table}

In case of test sets with shorter context, Table~\ref{NEDResultsOnShortDocs} shows that our best system cannot achieve better performance than its counterparts. The worst performed test sets compared to the SOTA results are KORE50 and RSS-500. KORE50 has the lowest average number of words per sentence and RSS-500 has only one mention per document on average. Our context-centric approach cannot utilize such short context enough. This is in fact the most common problem for all NED systems. On the other hand, Reuters-128 has a relatively larger context based on those metrics, however its average number of edit distance per candidate is the highest with respect to the other sets. That may cause the lower scores on Reuters-128. This is actually connected to the fact that AIDA.train on which we train our disambiguation model is characteristically different from these three data sets. One particular difference to mention is that the average edit distances given in Table~\ref{CandidateGeneratorRecalls} in the AIDA sets are lower than many of the other sets. Whenever our NED model gets  candidates with relatively higher edit distances on any of the test sets, it assigns low scores to such candidates as expected. This affects the gold candidates disproportionately, when there are alternative candidates with lower edit distances. 

It is interesting to note that the AQUAINT test set contains many concepts (\textit{e.g.,} ``power plant", ``radioactive waste" etc.) along with named entities. Our success might be related to the fact that we train our typing models on Wikipedia, which also includes the mentions of concepts.

\subsection{Analysis of the Experiments}

\subsubsection{Ablation Study on the Ranking Features}

In order to understand the contribution of the ranking features, we perform two ablation tests. Moreover, since we do the ranking in two stages, we calculate the contributions for both stages as well as for the baseline, which is the system that doesn't use any features obtained from the typing models. Before getting into the analysis, one factor has to be noted while evaluating the contributions. The second stage uses additional ``second stage" features and their success depends on the success of the features used in the first stage. Since those first stage features are still used in the second stage, their contribution drops as they share their success with the second stage features. Hence, it is more appropriate to evaluate the contribution of the first stage features based on the results at the first stage.

The first test examines the contribution of each feature by excluding that feature from the model. Table~\ref{FeatureAblationResults} lists all the features used in our experiments. Note that some of them are not available (N/A) for certain stages. The results\footnote{We train each model 10 time with different seed values and report the average results.} are shown in the F1-scores taken on the AIDA.testa (dev) set. As each feature is excluded, the drop in the F1 score is given. The first thing to notice is the high-level contribution of edit distance based features. Surface form is the major factor at the disambiguation task. The second thing is the decreasing contributions towards the second stage.

\begin{table}[h]
\begin{center}
\scriptsize
\renewcommand\arraystretch{1.0}
\begin{tabular}{lccc} \hline \hline 
\bf System                                                  & Baseline  & $1^{st}$ Stage & $2^{nd}$ Stage  \\ \hline
All Features Included                                       &  82.1  &  93.3   & 94.4     \\ \hline
\bf Local Features:                                         &        &         &        \\ 
- \textit{surface-form-edit-distance} (1)                   &  -9.7  &  -4.0   &  -1.3  \\ 
- \textit{entity-log-freq} (2)                              &  -3.0  &  -1.1   &  -0.1  \\
- \textit{surface-from-type-in-binary} (3)                  &  -2.4  &  -1.6   &  -0.6  \\ 
- \textit{entity-type-in-binary} (4)                        &  -1.8  &  -0.9   &  -0.4  \\
- \textit{$id^t$-typing-prob} (5)                           &  N/A   &  -0.2   &  -0.2  \\ \hline 
\bf Mention-level Features:                                 &        &         &        \\ 
- \textit{max-diff-surface-form-log-freq} (6)               &  -4.2  &  -0.8   &  -0.4  \\ 
- \textit{max-diff-doc-similarity} (7)                      &  -4.1  &  -0.6   &  -0.2  \\ 
- \textit{avg-surface-form-edit-distance} (8)               &  -0.6  &  -0.6   &  -0.3  \\
- \textit{max-diff-$id^t$-typing-prob} (9)                  &  N/A   &  -0.5   &  -0.3  \\  \hline
\bf Document-level Features:                                &        &         &        \\
- \textit{max-$id^t$-typing-prob-in-doc} (10)               &  N/A   &  -0.7   &  -0.2  \\ \hline 
\bf Second-stage Features:                                  &        &         &        \\
- \textit{max-ranking-score} (11)                           &  N/A   &  N/A    &  -0.5  \\ 
- \textit{max-cos-sim-in-context} (12)                      &  N/A   &  N/A    &  -0.8  \\ \hline \hline
\end{tabular}
\end{center}
\caption{\label{FeatureAblationResults}Showing the Contribution of each Feature in F1 Scores by its Exclusion at Different Stages on the AIDA.testa Development Set.}
\end{table}

Even though there are four groups of features shown in Table~\ref{FeatureAblationResults}, for the sake of contribution analysis, it is better to group them as surface form based (numbered 1,3,6,8), candidate based (2,7), typing model based (5,9,10), and ranking based (11,12) features. Each group considers the disambiguation task from different aspects.

\textbf{Surface form based features} are mention oriented, completely independent from the context. The results in Table~\ref{FeatureAblationResults} show that they keep their contribution high at all stages. This can be attributed to the fact that other features do not consider surface form. In other words, there is no contribution overlap between surface form based features and other features.

\textbf{Typing model features} are mostly context driven. Their contribution in Table~\ref{FeatureAblationResults} does not look notable at the individual level. However, the change between the Baseline and the $1^{st}$ stage, which is 11.2, comes from adding the typing model features. The low individual contributions can be explained by the fact that those three features are derived from the same value hence, their contributions overlap. Interestingly, using the highest typing model probability seen in other occurrences of the same candidate in the same document (10) is more powerful than using the typing model probability of the candidate (5) for the considered mention.

\textbf{Candidate based features} are more candidate specific, with no direct involvement of the surface form. Their contribution drops substantially at the 2$^nd$ stage. Even though (2) is widely used in the NED literature, it is calculated offline and independent of the mention or its context. This might explain its low performance. In case of (7), it is based on doc2vec embedding similarity between candidate's Wikipedia page and the test document. In other words, it is similar to the other typing model features. The results support that their contributions overlap.

\textbf{Ranking based features} help us take into account the surrounding candidates. The results show that they contribute quite well compared to others. Especially (12) might be the main driver behind the F1 score improvement of the 2$^{nd}$ stage.

In our further analysis, we observe that using the \textit{max-diff} version of the features performs better than using the feature itself. \textit{max-diff} compares the value of the feature with respect to the highest value seen among its sibling candidates. For example, using only \textit{doc-similarity} does not contribute better than \textit{max-diff-doc-similarity}. The same is true for \textit{max-diff-surface-form-log-freq}. We argue that this is related to the fact that \textit{doc-similarity} and \textit{surface-form-log-freq} are context independent and applying \textit{max-diff} makes them context dependent.

\begin{table}[h]
\begin{center}
\scriptsize
\renewcommand\arraystretch{1.1}
\begin{tabular}{lcc} \hline \hline 
\bf System                            & $1^{st}$ Stage & $2^{nd}$ Stage  \\ \hline
All Mention Typing Model Features Included    &  93.3 &  94.4   \\ \hline
- \textit{Word}                       &  -1.1 &  -0.6   \\ 
- \textit{Synset}                     &  -1.4 &  -0.7   \\
- \textit{Brown}                      &  -0.6 &  -0.2   \\
- \textit{Surface}                    &  -0.6 &  -0.3   \\ 
- \textit{Entity}                     &  N/A  &  -0.2   \\  \hline \hline
\end{tabular}
\end{center}
\caption{\label{TypingModelAblationResults}Showing the Contribution of each Mention Typing Model in F1 Scores by its Exclusion at Different Stages on the AIDA.testa Development Set.}
\end{table}

Our second ablation analysis evaluates the contribution of each typing model. Table~\ref{TypingModelAblationResults} shows the amount of drop in F1 score when we exclude each typing model from the $1^{st}$ and $2^{nd}$ stages on the AIDA.testa (dev) set. Note that features (5,9,10) shown in Table~\ref{FeatureAblationResults} have multiple values, one for each used typing model $t$. Considering that the \textit{Synset} typing model is the worst one at predicting the types as shown in Table~\ref{ModelResults}, it is surprising to see that it is the best contributing model. Like \textit{Synset}, the other local-context based model, namely the \textit{Word} typing model comes second. This indicates that typing models based on local (sentence-based) context contribute to the success of NED more than typing models based on global (document-based) context.

\subsubsection{Error Analysis of the Ranking Model}

In order to understand where our ranking model fails, we analyzed the failed cases on the AIDA.testa (dev) set in detail. Our first analysis involves measuring the role of the popularity of a candidate and the frequency of its surface form. It is intuitive to assume that popular entities are more likely to be mentioned than less popular entities. In our case, we define the popularity of a named entity as the total number of times that named entity is seen in the Surface Form Data Set. Similarly, when one surface form is seen more times with a certain entity than others, it is more likely that that surface form refers to that entity whenever it is used. When one surface form may refer to many entities, it causes ambiguity, which is the main source of the failures for the ranking model.

In Table~\ref{ErrorAnalysisTable}, we calculate certain ratios and percentages based on the popularity of the candidate ($F^e$) and the frequency of its surface form ($F^s$). During these calculations, $G$ is the set of the gold (true) candidates, one for each mention in the data set. $G'$ is its subset that only includes the ones (i.e. failed golds) that are failed to be predicted correctly by our model. $P'$ is the set of wrong predictions that are suggested instead of gold candidates. Note that $X$ is a variable in shown calculations and it should be replaced with either $F^e$ or $F^s$, when appropriate. Moreover, $X_{max}$ refers to the maximum value among all non-gold candidates of the same mention for the selected $X$.

\begin{table}[h]
\begin{center}
\scriptsize
\begin{tabular}{cllcc} \hline \hline 
\bf Case & \bf Description & \bf Calculation                                                   & $X=F^e$ & $X=F^s$  \\ \hline
[1]  & Ratio of failed golds' $X$ & $avg \sum_{g' \in G'} X_{g'} / avg \sum_{g \in G} X_{g}$       &  0.17  &  0.05   \\
     & to all golds' $X$ & & \\ \hline
[2]  & Ratio of predictions' $X$ & $avg \sum_{p' \in P'} X_{p'} / avg \sum_{g' \in G'} X_{g'} $      &  2.78  & 10.69   \\
     & to failed golds' $X$ & & \\ \hline
[3]  & \% of cases when gold's $X$ is & $\sum_{g \in G} [ X_{g} > X_{max} ] / size(G) $  &  0.36  & 0.48    \\
     & higher than the max's $X$ & & \\ \hline
[4]  & \% of cases when failed gold's  & $\sum_{g' \in G'} [ X_{g'} > X_{max} ] / size(G') $   &  0.29  & 0.16 \\ 
     & X is higher than the max's $X$ & & \\ \hline \hline
\end{tabular}
\end{center}
\caption{\label{ErrorAnalysisTable}Analyzing the impact of the popularity of the candidate entity and its surface form frequency when our system fails on the AIDA.testa (dev) set.}
\end{table}

Case [1] in Table~\ref{ErrorAnalysisTable} looks at the ratio of failed golds' popularity and form frequency to the all golds' in terms of averages. The values 0.17 for $F^e$ means that the average popularity of failed golds is about five times lower than the average popularity of the golds. In case of $F^s$, we can say that when our system fails, the form frequency of the gold candidate is 20 times less frequent than the average gold candidate's surface form.

In Case [2], we compare the values of wrongly predicted candidates and the corresponding gold candidates. For $F^e$, it shows that the popularity of the wrongly predicted candidate is 2.78 times higher than the popularity of the actual gold candidate. This is 10.69 times in case of $F^s$, which means that the surface form frequency plays more role than the popularity.

Case [3] measures the percentage of the cases when gold's popularity or form frequency is higher than any other sibling candidate. The nominator of [3] in Table~\ref{ErrorAnalysisTable} counts for how many $g$ $X_g$ is higher than $X_{max}$. The denominator normalizes that value to get the percentage. Case [4] measures the same value for the failed golds. In case of $F^e$, the value 0.36 means that 36\% of the time the golds are the most popular candidate and it is 29\% for the failed golds. In other words, most of the time golds are not the most popular ones among the candidates\footnote{This might be the reason why the \textit{entity-log-freq} feature does not have a notable impact on the disambiguation results.} and that ratio does not change much among the failed golds. In case of $F^s$, half of the time gold candidates are the ones that the surface form refers to most. That ratio drops considerably in case of failed golds. All in all, our system pays more attention to the form frequency than the popularity and prefers the candidates that the surface form refers to most. This also aligns with Case [2].

In addition to this analysis, we also measure the role of edit distance at failures. There are 15 gold candidates out of 4791 golds that have non-zero edit distance value in the AIDA.testa set. Meaning that, none of their surface form in our database matches with the surface form of the mention exactly. Our system fails 10 of those 15 cases. Compared to a total of 282 failures, it is less than 4\% of errors. However, in case of test sets like Reuters-128, it makes the difference.

Another aspect we analyze is how accurate our system predicts when there are multiple mentions of the same entity. Out of 4791 mentions, 2863 of them involve multi-mention cases. Our system fails at 153 of those cases, which does not look significant. Yet, it also means that more than half of the total failures of our system involve such cases. Moreover, in 60\% of 153 cases, our system predicts none of the instances of the multiple-times mentioned entity correctly.

\section{Discussion and Future Work}

In this study, we introduce a cluster-based mention typing approach. We cluster named entities based on their contextual embeddings and assign those cluster ids as type labels to entities. Our analysis shows that using a window as short as two to calculate entity embeddings with word2vec turns out to give better clusterings for the disambiguation task. We calculate five different clusterings of over five million named entities, each considering different contextual aspect. Based on these, we train five different mention typing models. The results show that LSTM-based models achieve better results than CNN-based models. Moreover, the models that use document-level context predict the cluster-based types better than the models with sentence-level context. We also introduce two specialized word embeddings that are influenced by the presence of cluster information during their calculation. Their analysis shows that such influence helps the typing model better predict the cluster-based types.

The second contribution of our study is to use the predictions of the mention typing model as features for the disambiguation of named entities. Our analysis shows that each typing model improves the disambiguation performance. However, using one typing model is not enough to achieve state-of-the-art (SOTA) results. As we use five of the models together, our system achieves better or comparable results based on randomization tests with respect to the state-of-the-art levels on four defacto test sets. Considering that our ranking model is a simple two-layer feedforward neural network, we score each candidate individually in a binary classification-based approach rather than employing a collective disambiguation approach. Moreover, we use the top 100 candidates rather than the top 20 and 30 as in previous works. Achieving SOTA results indicates the potential of using mention typing models. Our further analysis shows that even though the typing models with sentence-level context obtain lower scores at predicting the types, they are the most contributing models compared to document-level models at the disambiguation step.

Additionally, we study the candidate generation step. Our analysis shows that using candidates from similarly titled surrounding mentions and extending the candidates even further with their cooccurring entities from Wikipedia gives higher recall values.

There are a number of areas that can be addressed as future work to improve the proposed  system. One of the disadvantages of our mention typing model is its dependency on the context. The shorter it is, the worse it performs. One solution is to utilize existing context as much as possible with techniques like attention. Similarly, latest advances in embeddings such as BERT \cite{Devlin19} can be another alternative, as their context-customized embeddings help better represent the context. Another problem is the fact that our weakest typing model is the \textit{Synset}-based model. It can be argued that lack of contextual information in the \textit{Synset}-based types leads to a low performing typing model. Despite its high performance at the disambiguation step, finding a solution to this drawback might improve our weakest model and lead to even better results. Another improvement area is related to clustering. In our study, we obtain each clustering of the named entity space independently from the others and pick the best combination for the disambiguation step in Section~\ref{Sec:ClusterOptimization}. We face the dilemma of using larger cluster numbers to improve their discriminative power for the disambiguation step versus being less accurate at predicting the cluster-based types. These two contradicting criteria may lead to a not optimal solution. Nevertheless, we still achieve SOTA results. However, a more optimized clustering approach (such as customized K-means) that considers these two criteria together can produce better clusters. Customized clustering can even be designed to minimize the overlap between different types of clusterings. Alternatively, feedback from the disambiguation step can be circled back to the clustering step for iterative revision of clusters. One of the promising areas that our cluster-based types can help improve is NIL clustering, which is the task of clustering the mentions of NIL entities (i.e., entities that have no corresponding entries in the KB) and identifying all the mentions that correspond to the same NIL entity. In addition to using the surface forms of the mentions, utilizing their context is key to success. Prediction of the cluster-based types might provide extra clues for this task. This can be especially helpful at the corpus level (\textit{i.e.} cross-document) as there is more context to use to predict cluster-based types.

\section{Acknowledgements}

This research is supported by Bo\u{g}azi\c{c}i University Research Fund Grant Number 14201 and the Turkish Directorate of Strategy and Budget under the TAM Project Number 2007K12-873. Arda Celebi is also supported by an ASELSAN Graduate Scholarship for Turkish Academicians. GEBIP Award of the Turkish Academy of Sciences (to Arzucan Ozgur) is gratefully acknowledged.

\label{lastpage}

\end{document}